\documentclass{fairmeta}
\usepackage{amsmath}
\usepackage{enumerate} 
\usepackage{bm}
\usepackage{algorithm}
\usepackage{floatflt}
\usepackage{algpseudocode}
\usepackage{amsfonts}
\usepackage{amsthm}
\usepackage{newtxtt}
\usepackage{algorithm}
\usepackage{colortbl} 
\usepackage{cleveref}
\usepackage{diagbox}
\usepackage[utf8]{inputenc}
\usepackage{textgreek}
\usepackage{colortbl}
\usepackage{nicematrix}
\usepackage{makecell}
\usepackage{float}
\usepackage{arydshln}
\usepackage[frozencache,cachedir=.]{minted}
\usepackage[justification=centering]{caption}
\usepackage{subcaption}
\usepackage{tcolorbox}
\usepackage{amssymb}
\usepackage{xspace}
\usepackage{wrapfig}
\usepackage{adjustbox}
\usepackage{tabularx}
\usepackage{booktabs}
\usepackage{mathtools}
\usepackage{wrapfig}
\usepackage{amssymb}
\usepackage{graphicx}

\usepackage[T1]{fontenc}
\usepackage{lmodern}

\usepackage{silence}
\usepackage[table]{xcolor}
\usepackage{ulem}
\makeatletter
\patchcmd{\wrong@fontshape}{\@gobbletwo}{}{}{}
\makeatother
\newcommand{\model}[1]{\metadata[Model]{#1}}

\newtheorem{theorem}{Theorem}[]

\newtheorem{remark1}[theorem]{Remark}

\definecolor{upColor}{RGB}{17,138,21}
\definecolor{downColor}{RGB}{174,36,67}

\title{GN0: Toward a Unified Paradigm for Generation, Evaluation, and Policy Learning in Visual-Language Navigation}
\author[1\dagger\ddagger]{\text{Xinhai Li}}
\author[1,2\dagger~~]{\text{Xiaotao Zhang}}
\author[1,3\dagger~~]{\text{Yuehao Huang}}
\author[1\dagger~~]{\text{Jiankun Dong}}
\author[1,4\dagger~~]{\text{Tianhang Wang}}
\author[1,5\dagger~~]{\text{Sunyao Zhou}}
\author[1,4\dagger~~]{\text{Yunzi Wu}}
\author[1,6]{\text{Chengnuo Sun}}
\author[1]{\text{Yunfei Ge}}
\author[1]{\text{Qizhen Weng}}
\author[1]{\text{Chi Zhang}}

\author[1*]{\text{Chenjia Bai}}
\author[1*]{\text{Xuelong Li}}

\affiliation[1]{Institute of Artificial Intelligence, China Telecom}
\affiliation[2]{Shanghai Jiao Tong University}
\affiliation[3]{Zhejiang University}
\affiliation[4]{Tongji University}
\affiliation[5]{Fudan University}
\affiliation[6]{Jiangsu University}

\contribution{
\textsuperscript{$\dagger$}Equal Contributions
\quad
\textsuperscript{$\ddagger$}Project Leader
\quad
\textsuperscript{*}Corresponding Authors
}
\date{June 3, 2026}
\project{\url{https://telehuman-gn0.github.io.}}
\code{\url{https://github.com/TeleHuman/GN0}}
\model{\url{https://huggingface.co/TeleEmbodied/GN-BAE}}
\metadata[Correspondence to] {Xinhai Li (\email{lixh92@chinatelecom.cn}) Chenjia Bai (\email{baicj@chinatelecom.cn})}

\abstract{
Embodied navigation connects intelligent agents with the physical world and serves as a fundamental capability for achieving general robotic intelligence. However, the availability and quality of navigation data have long limited the generalization ability of Vision-and-Language Navigation (VLN) systems and their capacity to handle long-horizon tasks. To bridge this gap, we curate diverse 3D scene resources and develop an automated pipeline for large-scale navigation data generation, resulting in the \textbf{GN-Matrix} dataset. Building upon a 3D Gaussian Splatting (3DGS) rendering engine, we further introduce a high-fidelity simulation platform that supports interactive roaming and collision-aware navigation. Furthermore, we propose \textbf{GN-Bench}, the first benchmark to provide BEV-based evaluation, incorporating dynamic 3DGS avatars to establish new standards for human-robot interaction evaluation. 
To fully leverage the interactive nature of the simulator, we develop an RL-driven navigation foundation model, \textbf{Break and Establish (BAE)}, after supervised learning, DAgger exposes the model to rollout-induced states and recovery behaviors beyond idealized offline demonstrations, breaking the narrow expert-centric distribution inherited from supervision and providing a better starting point for downstream RL exploration. This unified VLN paradigm seamlessly integrates both map-based and map-free settings, enabling a broad spectrum of tasks including instruction following, human following, and goal navigation. 
To the best of our knowledge, this work is the first to formalize high-fidelity 3DGS-rendered Bird’s Eye View (BEV) representations as a compact and efficient memory mechanism, thereby unlocking the latent spatial reasoning capabilities of Vision-and-Language Models (VLMs). 
The dual-system architecture of \textbf{GN-BAE} ensures both flexibility and high-efficiency deployment in real-world scenarios. 
Extensive evaluations on GN-Bench and VLN-CE, encompassing both quantitative and qualitative analyses, demonstrate that the proposed GN0 significantly outperforms existing state-of-the-art VLN methods.
Overall, this work, GN-Matrix, presents a unified framework spanning data, simulation, and learning, offering new insights for advancing embodied navigation in both academic research and industrial applications.

}

\begin{document}

\maketitle

\section{Introduction}
Scene data has always been a critical challenge for Vision-Language Navigation (VLN)~\cite{anderson2018vision, Jain2019StayOT, krantz2020beyond} large models. The performance of VLN models heavily depends on the quality and fidelity of the scene data they interact with. However, traditional scene representation methods, such as meshes, panoramas, and video sequences, often have limitations in terms of detail rendering and interactivity. This directly impacts the robustness of VLN performance and leads to a domain gap in applications. 3DGS~\cite{kerbl3Dgaussians}  has recently garnered significant attention in 3D reconstruction, with broad applications in scene reconstruction~\cite{huang20242d, chen2024pgsr} and 3D asset generation~\cite{poole2022dreamfusion, li2023gaussiandiffusion, zhao2025hunyuan3d, xiang2025structured}, and has also inspired growing research in embodied intelligence~\cite{lou2025robo, li2024robogsim, qureshi2025splatsim, wang2025rethinking, lei2025gaussnav}. Rendering navigation environments with Gaussian Splatting (GS) offers a promising pathway to fully unlock the potential of vision-language models (VLMs). Unlike conventional mesh-based simulators, which often suffer from limited visual fidelity and domain gaps, GS enables photorealistic rendering with rich geometric and appearance details. Meanwhile, approaches that rely on real-world video sequences are constrained by limited coverage, poor controllability, and high data collection costs. Since VLMs are inherently pretrained on large-scale image data, GS provides a more compatible visual interface to transfer pretrained knowledge into embodied navigation. We propose a high-quality 3D scene dataset, GN-Matrix, based on open-source 3DGS scene assets~\cite{miao2026towards}, 3D scene generation~\cite{worldgrow2025}, self-capture reconstruction scenes~\cite{schoenberger2016vote} and commercially purchased scene assets. This dataset provides strong data support for the development of the field. Additionally, considering that interaction with and service to humans is the ultimate goal of robotics, we have incorporated dynamic 3DGS avatars into the scene, providing high-quality human-robot interaction data to support navigation tasks in dynamic environments~\cite{dong2025ha}. Based on heuristic search and LLM agents, we generated 47 million navigation data sequences in simulated environments, covering tasks such as instruction following, human following, and goal navigation. Each sample includes BEV~\cite{philion2020lift, yang2023bevformer} maps, first-person view images, and historical views. The action space consists of both discrete actions and continuous pixel-level trajectories, providing comprehensive data support for training navigation foundation models.

Based on the 3DGS rendering engine, we present an innovative simulation and evaluation platform. This platform supports advanced VLN navigation tasks, including roaming and collision detection, providing models with a more realistic and interactive environment to enhance the navigation experience. A major innovation of this platform is the introduction of dynamic 3DGS avatars benchmark, the first benchmark to provide BEV-based evaluation, offering new evaluation standards for human-robot interaction in VLN. This benchmark effectively evaluates a model's ability to interact with humans in dynamic, realistic environments, advancing the capabilities of VLN technology.

On the navigation model, we have trained a navigation base model GN-BAE based on GN-Matrix, which is compatible with both map-based~\cite{chen2024mapgpt} and map-free~\cite{cai2025navdp} approaches. This flexible design supports various tasks, such as instruction following, human following, and goal navigation. One of the key contributions of this work is the introduction of high-fidelity Bird’s Eye View (BEV) understanding in VLN, making the navigation process more intuitive and efficient. Our Vision-and-Language Model (VLM)~\cite{Qwen3-VL} base model is pre-trained on a large number of real images, while the rendered 3DGS images achieve realism comparable to real-world photographs, thereby truly unlocking the potential of VLMs. This innovation not only significantly improves VLN performance but also expands its application across various fields. GN-BAE is trained using reinforcement learning (RL) ~\cite{rafailov2024directpreferenceoptimizationlanguage, shao2024deepseekmathpushinglimitsmathematical} on GN-Bench, activating the spatial perception capabilities of VLMs, enhancing the model's obstacle avoidance abilities, and effectively addressing data distribution challenges. To promote real-world applications, our model adopts a dual-system architecture, ensuring both flexibility and efficiency. This architecture enables the system to adapt to different application scenarios and environmental requirements, providing support for rapid deployment in both industrial and academic settings.

\begin{figure}[t]
    \centering
    \includegraphics[width=\linewidth]{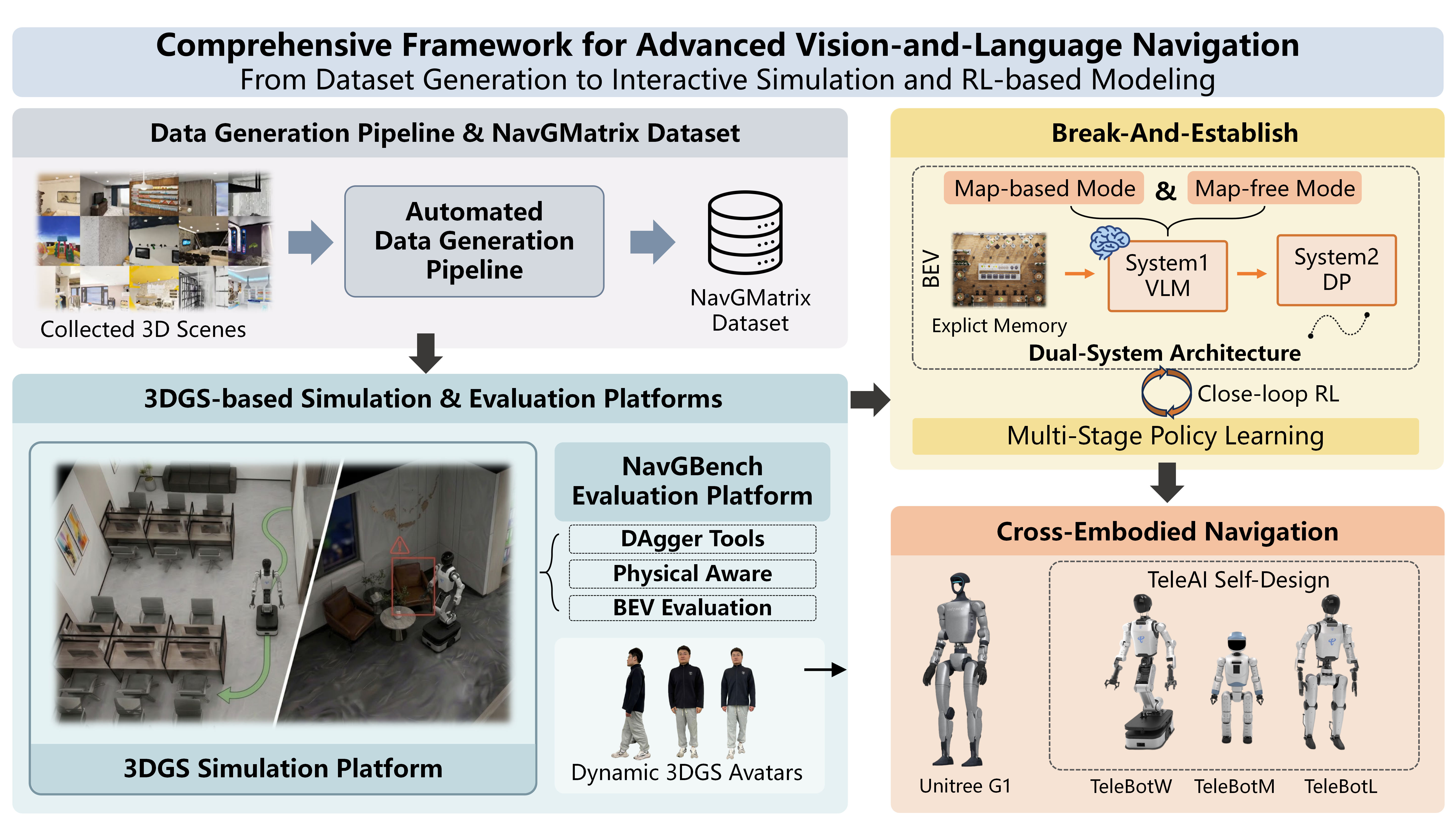}
    \caption{Overview of GN0, including GN-Matrix, GN-Bench, and GN-BAE.}
    \label{fig:main_overview}
\end{figure}

The seamless integration of GN-Matrix, GN-Bench, GN-BAE provides a comprehensive solution for advancing VLN technology. Our contributions are as follows:
\begin{itemize}

    \item A high-fidelity scene asset dataset, GN-Matrix, that integrates various indoor and outdoor assets, including open-source, commercial, and generated/reconstructed assets with dynamic 3DGS Avatars, along with a fully automated pipeline for generating navigation task data.

    \item A fully 3DGS-based navigation simulation platform and benchmark, GN-Bench, which is the first to incorporate BEV evaluation and introduces new evaluation standards for human-robot interaction in dynamic environments with moving agents, while supporting reinforcement learning-based exploration and learning for embodied agents with diverse embodiments within the scene.
    
    \item A navigation foundation model, GN-BAE,  unifies map-based and map-free paradigms, bridges academic research and industrial deployment. For the first time, high-fidelity 3DGS-rendered BEV representations are formalized as a compact memory mechanism, unlocking the latent spatial reasoning capabilities of VLMs and enabling diverse tasks such as instruction following, human following, and object-goal navigation. The dual-system architecture is optimized for high-throughput industrial inference, while a hybrid “Dagger + RL” training strategy first breaks the narrow expert distribution through DAgger and then enhances corrective exploration and continual learning via reinforcement learning. As a result, GN-BAE achieves state-of-the-art performance on both GN-Bench and VLN-CE benchmarks.

\end{itemize}

\section{Related Works}

\subsection{Vision-Language Navigation}

Embodied Vision-and-Language Navigation (VLN)~\cite{wang2025trackvla, cao2025cognav, gao2025octonav,liu2025trackvla++,zhou2026deconav} studies how agents follow natural-language instructions to move and act in complex environments, which is crucial for enabling embodied intelligence to generate practical value in real-world applications~\cite{an2026ai}. Existing methods are commonly developed under either discrete or continuous navigation settings. In discrete VLN~\cite{Chen2022ThinkGA, Zhou2023NavGPTER}, an agent moves over a predefined panoramic connectivity graph. This formulation simplifies high-level reasoning and route selection, but it also removes much of the metric structure, obstacle geometry, and collision constraints that are crucial for embodied deployment~\cite{xu2025mm}. Recent work~\cite{zheng2024towards, huang2025cogddn} has further incorporated LLM-based planners and retrieval-augmented memory modules to improve instruction understanding, history utilization, and subgoal decomposition. Nevertheless, these approaches often depend on oracle graph connectivity and provide limited verification of whether the generated plans are physically executable at the low-control level. Continuous VLN aims to close this gap by requiring agents to act directly in metric environments. A common strategy is to combine high-level reasoning with pretrained waypoint prediction, but such predictors can overfit to training environments and may generalize poorly to unseen layouts or visual conditions. To reduce dependence on scene-specific priors, recent end-to-end dual-system architectures have been explored~\cite{internvla-n1,chu2026abotn0technicalreportvla}. These methods typically use a high-level module for embodied reasoning with slow/fast contextual memory, together with a low-level controller, such as a diffusion-policy-based motion module, for responsive local action execution. In this work, we introduce high-fidelity bird's-eye-view (BEV) representations to construct dual-system memory, providing geometrically grounded spatial context for both long-term reasoning and short-term control, thereby improving navigation robustness in dynamic embodied environments.

\subsection{Scene Representations for Vision-Language Navigation}
Effective spatial memory plays a central role in enabling embodied VLN agents to perceive, organize, and reason about their surroundings. Existing studies have investigated this problem from different perspectives. ETPNav~\cite{an2024etpnav} maintains memory through evolvable scene graphs, which support long-horizon navigation by continuously updating topological and semantic relations in the environment. Moving beyond purely topological representations, FOM-Nav~\cite{chabal2025fomnavfrontierobjectmapsobject} and 3D-Mem~\cite{yang20253d} employ vision-language models (VLMs) to construct and update high-level semantic map memories. By exploiting the zero-shot reasoning ability of VLMs, these methods aim to preserve object-level consistency across observations from different viewpoints.
Another line of work investigates implicit memory learning. Uni-NaVid~\cite{zhang2024uni}, for example, adopts an end-to-end video-based architecture to model both short-term and long-term navigation history within latent representations. In contrast to these graph-based, semantic-map-based, or implicit-memory formulations, our method builds an explicit spatial memory using bird's-eye-view (BEV) representations rendered from 3D Gaussian Splatting~\cite{kerbl3Dgaussians}. This design provides a geometrically grounded memory structure that strengthens both global scene-level spatial awareness and ego-centric local understanding.

\subsection{Language-driven Navigation Benchmarks}

Recent progress in Vision-and-Language Navigation (VLN) has been closely tied to the development of benchmarks that progressively increase task difficulty, environmental realism, and evaluation fidelity.
Early datasets such as Room-to-Room (R2R)~\cite{anderson2018vision} and Room-for-Room (R4R)~\cite{Jain2019StayOT} formulate VLN as instruction following over predefined panoramic paths, where agents are required to execute step-by-step navigation commands in static indoor scenes.
Subsequent efforts move beyond discrete panoramic graphs. VLN-CE~\cite{anderson2018vision} introduces continuous control in photorealistic simulation environments, placing greater emphasis on visual perception and fine-grained action execution.
More recently, VLN-PE~\cite{wang2025rethinking} further improves physical realism by modeling robot dynamics and control imperfections that commonly arise during real-world deployment.
Beyond the standard static-environment assumption, several recent benchmarks have begun to explore more complex and interactive navigation settings.
HA-VLN~\cite{dong2025ha} incorporates human activities into VLN scenarios, enabling agents to navigate in environments with dynamic human motion.
SocNav~\cite{chen2026socialnavtraininghumaninspiredfoundation} focuses on socially aware navigation by providing diverse social scenes and a comprehensive set of metrics for evaluating human-aware behaviors.
In parallel, SAGE-Bench~\cite{miao2026towards}, built upon InteriorGS~\cite{InteriorGS2025}, leverages 3D Gaussian Splatting (3DGS)~\cite{kerbl3Dgaussians} to transform high-fidelity scene representations from passive perceptual assets into executable environments with semantic and physical alignment for embodied navigation.
Other recent benchmarks further expand the scope of VLN toward long-horizon instruction following~\cite{song2025towards} and collaborative embodied navigation~\cite{wang2026conavbench}.
Beyond standard VLN benchmarks, recent advances in multimodal spatial reasoning provide inspiration for embodied navigation. Models such as AssemLM ~\cite{jing2026assemlm} demonstrate that integrating point clouds, textual instructions, and visual manuals enables precise 6D pose reasoning and generalization across diverse object categories in robotic assembly tasks. 
Despite these advances, existing benchmarks still offer limited support for jointly studying 3DGS-based simulation, dynamic agents, and human-robot interaction within a unified navigation framework.
To address this gap, we propose GN-Bench, a fully 3DGS-based navigation simulation platform and benchmark designed for human-robot interactive navigation in dynamic environments with moving agents.

\section{Data Generation}
\label{sec:data_generation}

\subsection{Overview and Key Characteristics}
\label{subsec:datagen_overview}

\begin{figure}[t]
    \centering
    \includegraphics[width=\linewidth]{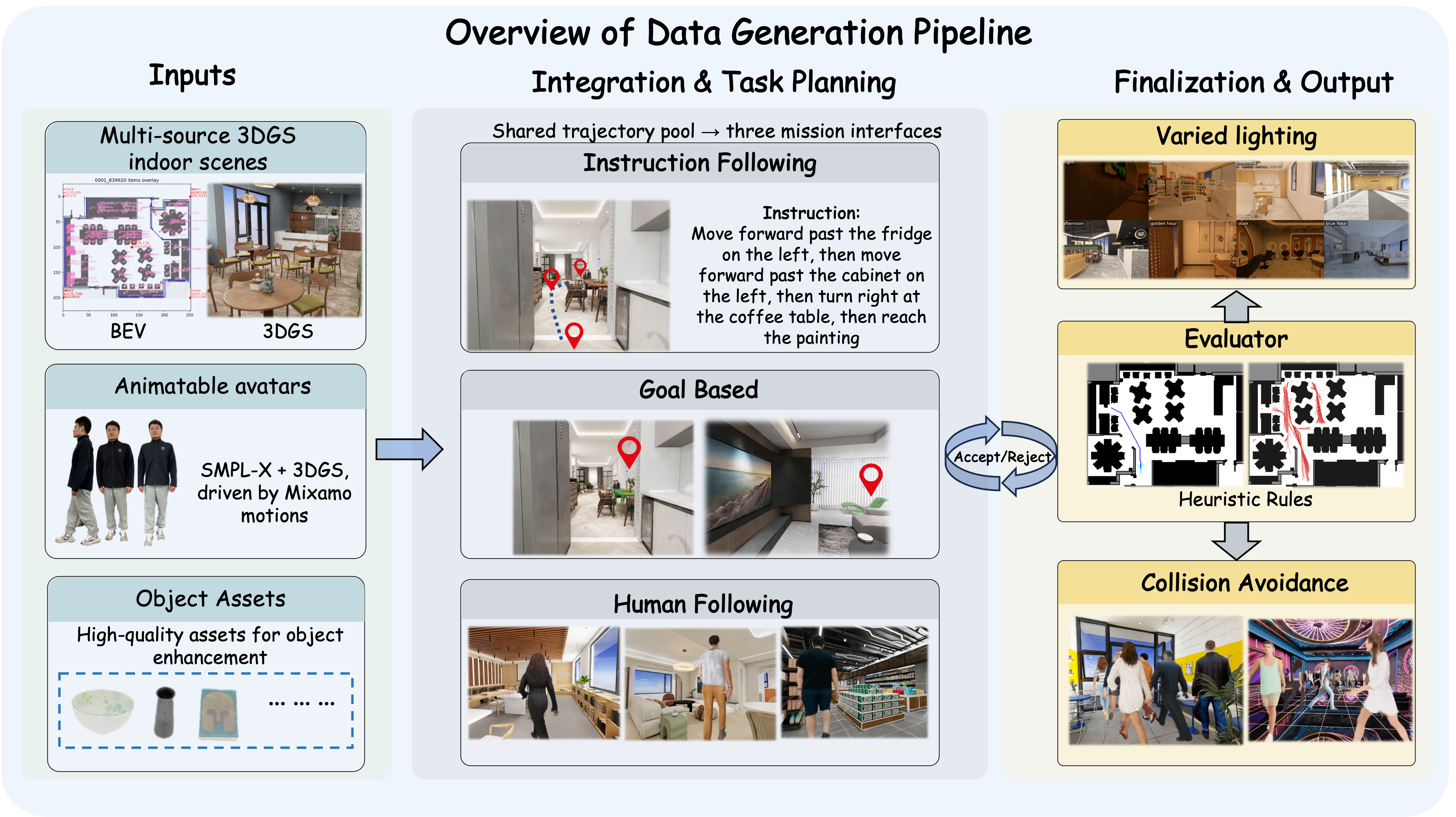}
    \caption{Overview of the GN-Matrix data generation pipeline.}
    \label{fig:data_pipeline}
\end{figure}

To support large-scale embodied navigation in photorealistic
3D Gaussian Splatting (3DGS) environments, we construct
\textbf{GN-Matrix}, a large-scale navigation dataset covering diverse
indoor scenes, dynamic human-centered environments, and
benchmark-compatible data formats for multiple navigation settings
\cite{zhang2024navid,krantz2020beyond,wang2025rethinking}. To build
GN-Matrix, we develop a unified data generation pipeline that
integrates multi-source 3DGS scene assets, animated 3DGS human
avatars, task synthesis, and multimodal annotation. As illustrated in
Fig.~\ref{fig:data_pipeline}, this pipeline is designed not only to
increase the scale of navigation data, but also to improve scene
diversity, task coverage, and realism.

The key characteristics of GN-Matrix are summarized as follows:
\begin{itemize}
    \item \textbf{High-quality 3DGS scene foundation.} GN-Matrix is
    built on visually faithful 3DGS scene representations, providing
    realistic observations for embodied navigation in indoor
    environments.

    \item \textbf{Diverse scene assets.} The dataset combines
    InteriorGS, handcrafted large-scale scenes, and WorldGrow-expanded
    scenes to improve scene diversity, layout complexity, and
    navigational difficulty.

    \item \textbf{Dynamic human avatars.} Beyond static environments,
    GN-Matrix includes animated 3DGS human avatars with diverse
    appearances and motions, enabling human-centered embodied
    navigation settings.

    \item \textbf{Three complementary mission types.} GN-Matrix supports
    three navigation mission interfaces built on top of a shared
    trajectory-generation backbone: goal-based navigation,
    instruction-following navigation, and human-following navigation. These
    mission types can reuse the same feasible start--goal paths while
    exposing different forms of task specification, including visual goal
    references, route-level language specifications, and dynamic human
    targets.

    \item \textbf{Multimodal and benchmark-compatible supervision.}
    The dataset provides multimodal annotations and can be adapted to
    multiple downstream benchmark and training settings, including
    VLN-CE, VLN-PE, and NaVid.
\end{itemize}

Within the overall framework of this work, GN-Matrix provides the data
foundation, GN-Bench provides the benchmark and simulation
environment, and GN-BAE serves as the model baseline.

\subsection{Scene Assets}
\label{subsec:scene_assets}

GN-Matrix is built from multiple complementary 3DGS scene sources.
Rather than relying on a single dataset, we combine curated indoor
reconstructions, handcrafted large-scale scenes, and automatically
expanded environments to improve scene diversity, layout complexity,
and coverage of navigation difficulty. This multi-source design allows
the dataset to provide both high-quality visual grounding and broader
coverage of embodied navigation settings.

\paragraph{Unified scene representation.}
Although these scene sources differ in origin and construction method,
they are converted into a unified navigation-ready representation
before downstream data generation. This standardization ensures
consistent traversability modeling, semantic grounding, rendering
interfaces, and coordinate conventions across all scene types, making
it possible to apply a shared task synthesis and annotation pipeline
throughout the full dataset.

\subsubsection{InteriorGS}
\label{subsubsec:interiorgs}

We take the 1,000 scenes provided by InteriorGS~\cite{InteriorGS2025}
as the main foundation of GN-Matrix. InteriorGS provides realistic
indoor 3DGS reconstructions together with semantic annotations, making
it a strong starting point in terms of scene quality, visual realism,
and object-level grounding. These properties are particularly
important for navigation tasks that depend on visually grounded target
identification and scene understanding.

At the same time, although InteriorGS provides diverse and
well-annotated scenes, its average scene size and layout complexity
remain limited for long-horizon embodied navigation. This motivates
the inclusion of additional scene sources that emphasize larger-scale
spaces and more challenging navigable structures.

\subsubsection{Handcrafted Large-Scale Scenes}
\label{subsubsec:handcrafted_scenes}

To complement the InteriorGS base, we introduce 150 handcrafted,
dense, large-scale scenes covering environments such as offices and
supermarkets. Compared with typical indoor scans, these scenes provide
larger navigable areas, more complex spatial layouts, and richer
long-range navigation cases.

The main role of this scene source is to expand the dataset toward
navigation settings that require longer trajectories, more complex
layout reasoning, and denser environmental structure. As a result, the
handcrafted scenes strengthen GN-Matrix not only in scale, but also
in task difficulty and spatial diversity.

\subsubsection{WorldGrow-Expanded Scenes}
\label{subsubsec:worldgrow_scenes}

To further increase scene diversity and overall dataset scale, we also
adopt the WorldGrow~\cite{worldgrow2025} pipeline to expand the scene
set. In contrast to curated and handcrafted scene sources, this
component allows us to efficiently generate additional environments and
thereby broaden the distribution of indoor layouts, semantics, and
navigational conditions represented in the dataset.

Our WorldGrow-based scene expansion follows a structured generation
pipeline. First, based on the 3D-FRONT dataset, we develop a scene
chunking strategy to extract high-quality and spatially dense scene
blocks for training. Second, we improve the structured latent
representation in TRELLIS~\cite{xiang2025structured} by introducing an occlusion-aware feature
aggregation mechanism and retraining the decoder on scene data, in order
to mitigate object occlusion issues and boundary artifacts. Third, we
perform 3D block-based inpainting using a flow-matching Transformer,
which autoregressively generates adjacent scene blocks conditioned on
noise, masks, and known neighboring regions. Through this process, we
obtain expanded indoor scenes with improved scale and layout diversity.

The inclusion of WorldGrow-expanded scenes improves the scalability of
GN-Matrix and helps reduce over-reliance on a fixed set of manually
constructed environments. This is particularly useful for large-scale
data generation, where scene diversity is a key factor in improving
generalization.

\paragraph{Object quality improvement.}
To further improve the usability of WorldGrow-expanded scenes, we
construct an object enhancement pipeline based on voxel querying and
asset replacement. This enables us to replace repetitive, low-quality,
or visually corrupted generated objects with higher-quality and more
diverse assets derived from HSSD~\cite{khanna2023hssd}. Specifically,
we first convert selected HSSD assets into 3D Gaussian representations
through a pipeline that renders multi-view images in Blender and then
reconstructs Gaussian assets using Gaussian
Splatting~\cite{kerbl3Dgaussians}. These assets are then used to replace
low-fidelity furniture instances in the generated scenes. This process
addresses several shortcomings of WorldGrow-expanded environments,
including limited object diversity in color, texture, and structural
detail, as well as occasional deformed, warped, or otherwise corrupted
objects. As a result, it improves both the visual fidelity and semantic
density of the generated scenes, making them more suitable for embodied
navigation data generation.

\subsection{Dynamic Human Avatars}
\label{subsec:human_avatars}

A key feature that distinguishes GN-Matrix from purely static
navigation datasets is the inclusion of dynamic human avatars. Since
human-centered interaction is a central objective of embodied
intelligence, a navigation dataset should not be limited to empty or
static environments. Instead, the agent should operate in scenes that
contain diverse human appearances and motions, enabling more realistic
evaluation settings such as human-following, dynamic obstacle
avoidance, and navigation under time-varying scene conditions.

To construct this component, we build a human avatar bank from
single-view human images collected from SHHQ and HuGe100K, supplemented with generated human avatars. For each
input image, we recover a parametric SMPL-X body model using
SMPLify-X, while separately reconstructing a canonical 3D Gaussian
avatar in T-pose using LHM. These two representations provide
complementary advantages: the SMPL-X model offers a structured and
animatable body prior, while the reconstructed Gaussian avatar
preserves the detailed visual appearance of the person.

We then align the SMPL-X mesh to the canonical Gaussian avatar and
establish a binding relationship between the body model and the 3DGS
representation. This step allows the reconstructed Gaussian avatar to
be driven by body motion while preserving its appearance. In our
current pipeline, we consider two fitting strategies for this
alignment: an ICP-style nearest-neighbor correspondence method and a
robust statistical fitting formulation based on Mahalanobis distance.
The detailed mathematical formulations are provided in
Appendix~\ref{app:human_avatar_details}.

To animate the avatars, we first collect motion assets in FBX format
from Mixamo. Each animation sequence is converted into frame-wise
SMPL-X parameters, represented as per-frame \texttt{smplx.json} files.
These frame-wise body parameters are then used to drive the aligned
SMPL-X--Gaussian pair, producing a sequence of animated per-frame
Gaussian avatars. This process enables us to combine diverse human
appearances with a reusable motion library, thereby constructing
dynamic human representations suitable for embodied navigation in
human-centered environments.

Overall, the dynamic human avatar component allows GN-Matrix to move
beyond static scene navigation and support richer embodied settings in
which humans act as moving targets, dynamic scene elements, and future
interaction anchors.

\begin{figure}[h]
    \centering
    \includegraphics[width=\linewidth]{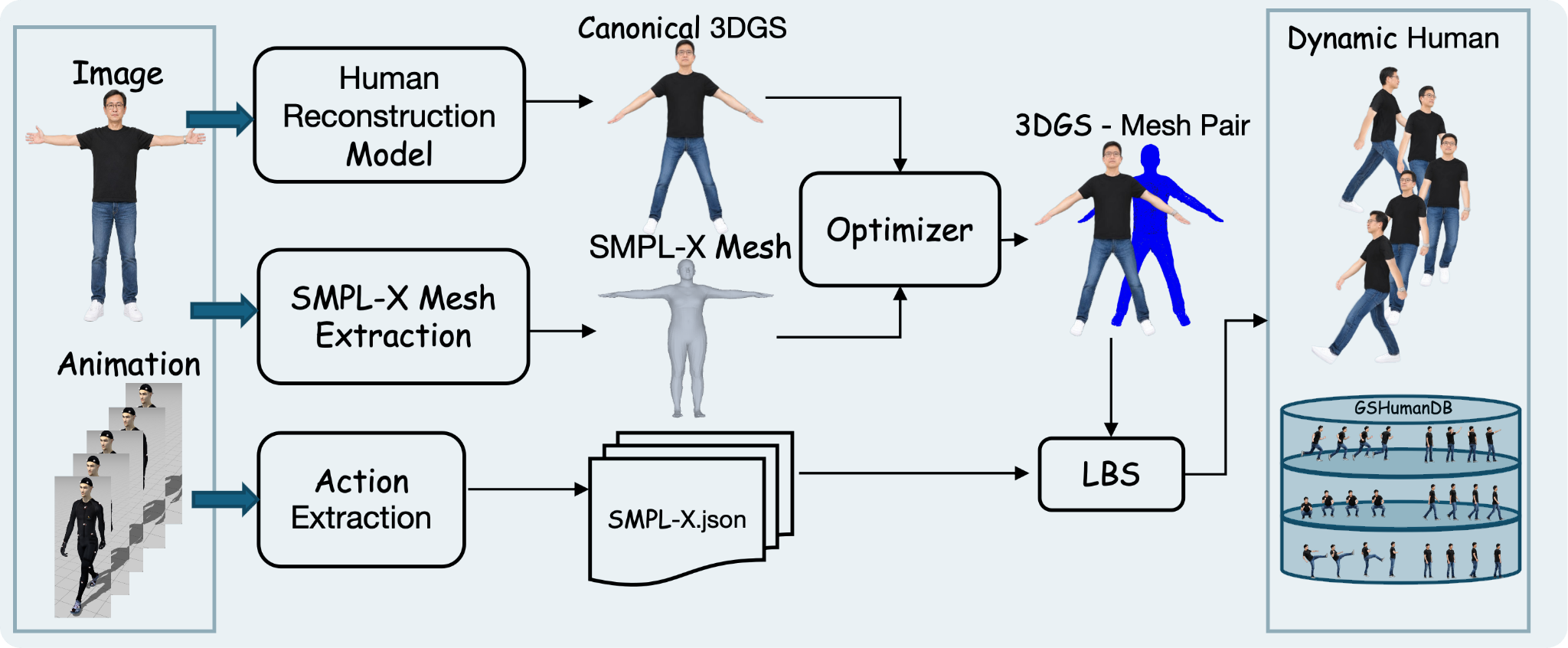}
    \caption{Pipeline for constructing animatable dynamic human avatars,
    including single-image reconstruction, mesh--Gaussian alignment, and
    motion-driven avatar generation.}
    \label{fig:human_datagen_pipeline}
\end{figure}
\subsection{Mission Types and Task Synthesis}
\label{subsec:mission_task_synthesis}
GN-Matrix separates trajectory generation from mission presentation. We
first construct a shared pool of feasible start--goal trajectories over
the traversable regions of each scene, and then instantiate each
accepted path under different mission interfaces. This design allows the
same spatial path to support multiple forms of supervision: a visual
object or region target, a detailed route instruction, or a dynamic
human target. As a result, the dataset can evaluate whether navigation
agents can follow the same underlying route when the task is specified
through different sensory and linguistic interfaces.

For each scene, we begin by collecting candidate target objects and
regions from the available semantic scene annotations to form a target
pool. Since raw scene annotations may contain multiple nearby instances
or repeated targets that do not meaningfully increase navigation
diversity, we further filter the target pool to suppress spatially
redundant or overlapping endpoints. This ensures that the generated
tasks are centered on meaningful navigation goals rather than densely
duplicated endpoint choices.

Given the filtered target pool, we sample feasible start--goal pairs
over the traversable region of the scene. The goal is not merely to
generate as many trajectories as possible, but to construct diverse and
non-trivial navigation episodes that cover different spatial regions,
path lengths, and scene configurations. To this end, start locations are
sampled from valid navigable areas, while goal locations are grounded in
semantic targets or selected endpoint regions. For each feasible pair,
we then generate a reference navigation trajectory over the scene
traversability structure.

To improve dataset quality, we further apply validity and diversity
constraints to the generated trajectories. In particular, we remove
episodes that are excessively short, spatially trivial, or highly
similar to previously accepted paths. This filtering step encourages the
final dataset to cover broader navigation patterns and reduces the
over-representation of near-duplicate trajectories. Each accepted
trajectory is then packaged with reference path information, waypoint or
action supervision, and the metadata required to instantiate the three
mission types described below.

\subsubsection{Goal-Based Navigation Tasks}
\label{subsubsec:goal_nav}

We first instantiate each accepted trajectory as a goal-based navigation
task. In this setting, the mission specifies the destination through a
rendered goal reference or target observation, together with a concise
target-oriented instruction. The agent is required to ground the target
within the 3DGS scene and navigate to the corresponding endpoint.

Unlike instruction-following navigation, this mission type specifies the
destination without explicitly prescribing the intermediate route. Each
episode is packaged with the goal reference, target semantics, start
pose, reference trajectory, and downstream waypoint or action
supervision. These generated tasks can later be converted into
multimodal training instances with first-person observations,
bird's-eye-view context, action or waypoint supervision, and
benchmark-compatible formats.

\subsubsection{Instruction-Following Navigation Tasks}
\label{subsubsec:instruction_nav}

The second mission type is instruction-following navigation. In this
setting, the underlying start--goal path can be identical to the
goal-based trajectory, but the task specification is provided as a
route-level natural-language instruction. Rather than specifying only
the destination, the instruction describes intermediate navigation
constraints, including landmark references, relative directions, and
ordered spatial decisions along the path.

This setting corresponds to the route-following interface used in many
vision-language navigation benchmarks. The instruction provides
intermediate constraints along the path, requiring the agent to align
language with local observations, maintain progress over a long-horizon
route, and execute the navigation sequence in the correct order.

To generate these tasks, we convert each accepted reference trajectory
into a route-level language specification using the scene semantics,
landmark visibility, turning points, and endpoint context along the
path. The resulting episode shares the same spatial trajectory as the
corresponding goal-based task, but exposes a different input interface.
This decoupling allows GN-Matrix to compare goal-based navigation and
language-guided route following under matched path geometry and scene
difficulty.

Each instruction-following episode is packaged with the route
instruction, start pose, endpoint metadata, reference trajectory, and
downstream supervision signals. These tasks support
instruction-conditioned training and evaluation while preserving the
same 3DGS grounding and action-space conventions used by the rest of
the dataset.

\subsubsection{Human-Following Tasks}
\label{subsubsec:human_follow}

Beyond traditional static goal-based navigation and instruction-following
navigation, GN-Matrix also supports human-following tasks in 
which the agent is required to follow 
a moving human target through the scene. Compared with the other two types of 
navigation, the target in this setting is no longer a fixed semantic
endpoint, but a dynamic human whose position evolves over time. This
task family therefore provides a more realistic testbed for
human-centered embodied navigation in dynamic environments.

To construct these tasks, we instantiate animated human avatars inside
the scene and assign them motion trajectories derived from the dynamic
avatar pipeline described in
Section~\ref{subsec:human_avatars}. Rather than generating an entirely
independent trajectory source, we adapt trajectories from the shared trajectory pool and reinterpret them as human motion
paths. In this way, the dynamic human target inherits feasible and
scene-consistent navigation behavior while extending the dataset from
static semantic goals to moving human-centered targets.

Given a moving human trajectory, we then construct a corresponding
agent-follow episode in which the agent navigates relative to the target
motion rather than toward a fixed endpoint. In particular, the task is
designed to preserve a reasonable following relationship, such as
maintaining a preferred relative distance and viewpoint with respect to
the moving target, while still adapting to scene geometry and
time-varying occlusion. This transforms the shared reference trajectory into a dynamic
navigation problem with continuous target tracking.

Compared with static goal-based and instruction-following navigation, 
human-following introduces additional challenges including motion-aware path
adjustment, changing target visibility, and the need to remain robust
to dynamic scene configurations. These properties make it a useful
benchmark setting for evaluating embodied navigation systems in more
realistic human-centered environments.

Each generated human-following episode is packaged with the target human
identity, the underlying motion trajectory, the associated agent
trajectory, and downstream supervision signals. These tasks can
therefore support multimodal training and evaluation for dynamic
navigation, while also serving as a foundation for future extensions
toward richer human-agent interaction scenarios.

\subsection{Benchmark Compatibility and Dataset Statistics}
\label{subsec:compatibility_stats}

\begin{figure*}[t]
    \centering
    \includegraphics[width=0.9\textwidth]{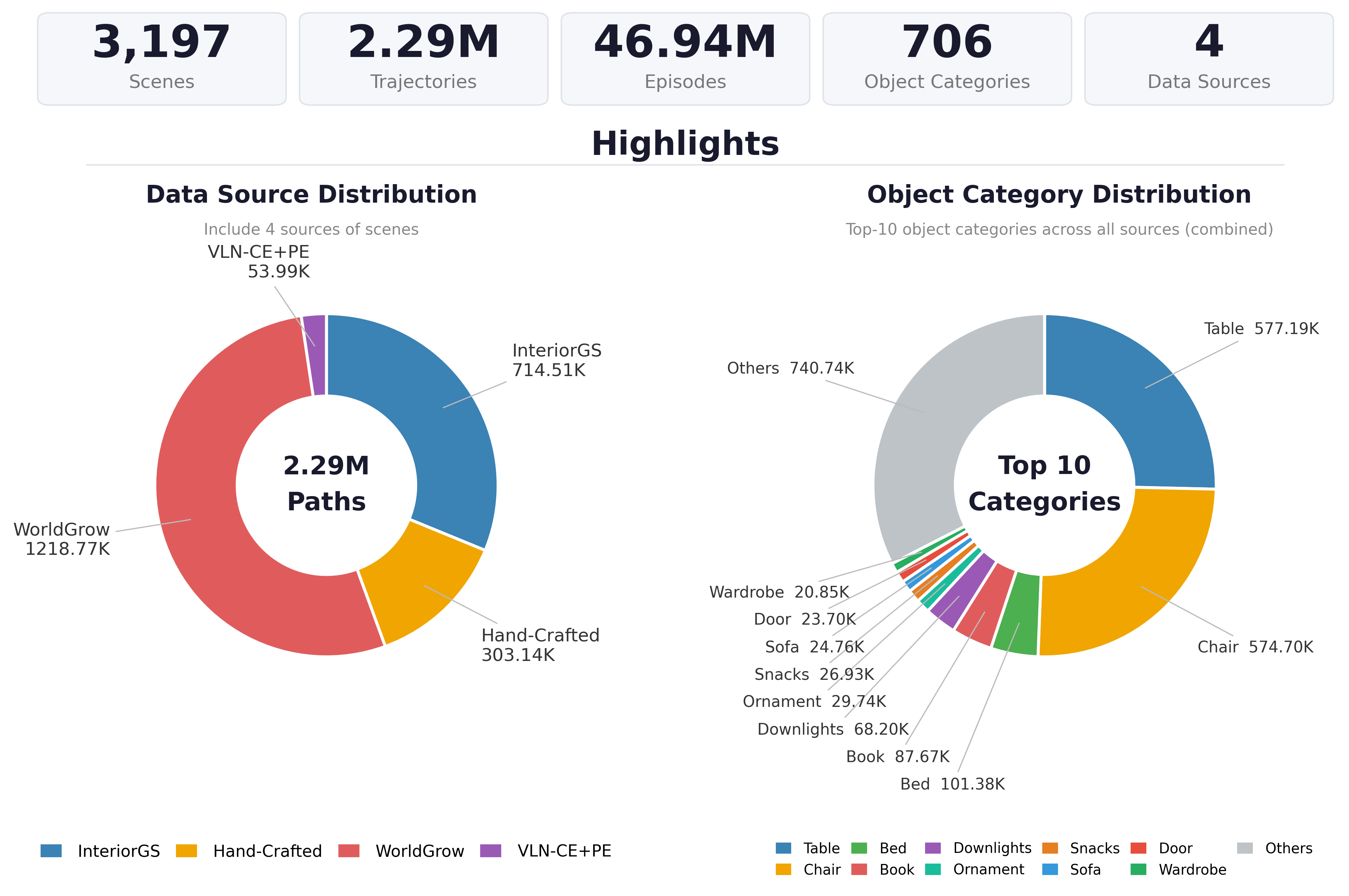}
    \caption{
    Dataset statistics of GN-Matrix. We summarize the overall dataset scale and composition of the generated navigation data, including the number of scenes, trajectories, episodes, object categories, data sources, source distribution, and object-category distribution.
    }
    \label{fig:GN-Matrix_stats}
\end{figure*}

In addition to serving as a native 3DGS-based embodied navigation
dataset, GN-Matrix is also designed to support compatibility with
existing embodied navigation benchmarks and training settings. Rather
than treating these benchmarks as separate data sources, we view them as
target export formats built on top of the same scene-grounded task
generation pipeline. This design allows GN-Matrix to preserve its
original spatial and semantic grounding while remaining usable within
established embodied navigation ecosystems.

To improve interoperability with existing benchmarks, we adapt the
generated mission specifications into benchmark-compatible formats.
Depending on the target setting, goal-based, instruction-following,
and human-following missions can be reformulated into
benchmark-compatible interfaces aligned with VLN-CE, VLN-PE, and NaVid,
while preserving the underlying scene-grounded trajectory and navigation
semantics.

Beyond instruction adaptation, we also convert the generated reference
trajectories into benchmark-compatible discrete action sequences. This
step is important because many embodied navigation benchmarks and policy
learning frameworks operate on discrete control spaces rather than
continuous trajectory representations. By aligning the generated task
data with this control convention, GN-Matrix can be used more directly
for downstream training and evaluation in standard navigation setups.

Using the above pipeline, we construct a large-scale navigation dataset
covering both static and dynamic embodied navigation tasks. GN-Matrix
contains multi-source indoor scenes, dynamic human-centered episodes,
and diverse mission types including goal-based navigation,
instruction-following navigation, and human-following navigation. We
further rebalance the raw action distribution to reduce the dominance
of forward actions and improve the usefulness of the dataset for policy
learning. The final dataset scale, mission composition, and
train/validation/test split are summarized in the main paper, while
detailed conversion rules and benchmark-specific export settings are
provided in the appendix.

\begin{table}[h]
    \caption{Summary of benchmark compatibility and dataset outputs.}
    \label{tab:compatibility_stats}
    \centering
    \small
    \begin{tabularx}{\textwidth}{l X l X}
    \toprule
    \textbf{Setting} & \textbf{Instruction Format} & \textbf{Action Space} & \textbf{Notes} \\
    \midrule
    Native GN-Matrix & Goal / instruction / human-following mission formats & Path / action supervision & 3DGS-based representation \\
    \midrule
    VLN-CE / VLN-PE / NaVid & Adapted & Discrete & Benchmark-compatible export \\
    \bottomrule
    \end{tabularx}
\end{table}

\subsection{DAgger Data Collection}
\label{sec:dagger}
To mitigate compounding errors caused by the covariate shift between offline expert demonstrations and online policy execution, we construct an exploratory dataset utilizing the Dataset Aggregation (DAgger) algorithm. During this collection phase, the agent actively explores the environments driven by the initial SFT policy. At each visited state $s_t$, the system dynamically relabels the current observation with optimal expert annotations for both continuous waypoints and discrete actions.

For goal-based and instruction-following missions, the oracle planner
uses the static endpoint as the destination. For human-following
missions, the oracle instead tracks the time-indexed human trajectory
and generates corrective labels relative to the desired following state.

For the continuous pixel trajectory, we query the integrated oracle
planner. By executing the A* search algorithm~\cite{4082128} over the
dilated occupancy grid $\mathcal{M}_{dil}$, we extract the optimal,
collision-free geodesic path from the current planning state to the
destination $g$. This global path is subsequently projected into the
agent's egocentric frame to formulate the ground-truth BEV waypoint
sequence $\mathcal{T}^*_{bev, t}$.

Concurrently, to generate the discrete action labels, we formulate a Model Predictive Control (MPC) strategy~\cite{RICHALET1978413}. The MPC optimizes a short-horizon control sequence to track the A* reference path, yielding the optimal VLN-CE compliant macro-action $a^*_t \in \{\text{\texttt{<FWD>}}, \text{\texttt{<LEFT>}}, \text{\texttt{<RIGHT>}}, \text{\texttt{<STOP>}}\}$. This expert action selection is formalized as minimizing the tracking cost over a predictive horizon $H$:

\begin{equation}
    \label{eq:mpc_action}
    a^*_t = \arg\min_{a \in \mathcal{A}} \sum_{k=1}^{H} \gamma^{k} \mathcal{C}(s_{t+k}, \mathcal{T}^*_{bev, t})
\end{equation}

where $\gamma$ is a discount factor and $\mathcal{C}$ denotes the spatial deviation cost from the optimal A* trajectory. 

Furthermore, to prevent the agent from infinitely hovering near the target without asserting task completion, we enforce a strict halting heuristic. Let $d(p_t, g)$ denote the spatial distance between the agent's current coordinate $p_t$ and the goal $g$, and let $d_{success}$ define the task success radius. Upon entering the success region ($d(p_t, g) \le d_{success}$), the system initializes a localized step counter $c_{goal}$. If the agent executes 10 additional steps without emitting the termination token, the simulation environment forcibly truncates the episode. Under this condition, the expert label for the terminal state is explicitly overwritten to supervise the halting behavior:

\begin{equation}
    \label{eq:halting}
    a^*_t = \text{\texttt{<STOP>}}, \quad \text{if } d(p_t, g) \le d_{success} \text{ and } c_{goal} \ge 10
\end{equation}

This aggregated dataset of on-policy states coupled with oracle-corrected targets provides the necessary corrective supervision to expand the policy space, directly supporting the entropy increase phase prior to GRPO training.

\subsection{Post-processing and Dataset Balancing}
\label{subsec:postprocess_balancing}

After task synthesis, we further improve the quality and usability of
GN-Matrix through post-processing and data balancing. These steps are
designed to increase both visual diversity and learning utility in the
final dataset.

\paragraph{Visual-condition enhancement.}
To improve robustness to appearance variation, we construct an
enhancement pipeline that generates additional data under diverse
lighting conditions by manipulating illumination intensity and color
schemes. This process produces visually varied observations while
preserving the underlying navigation semantics of each task, thereby
broadening the visual coverage of the final dataset.

\paragraph{Action distribution balancing.}
The raw generated trajectories are strongly biased toward forward
actions, with comparatively fewer turning and stopping actions. To
reduce this imbalance, we rebalance the final dataset by adjusting the
relative proportions of forward, turning, and stop actions. This
improves the usefulness of the dataset for downstream navigation policy
learning and reduces the tendency of models to over-predict straight
motion.

\section{GN-Bench}
GN-Bench is a high-fidelity embodied navigation benchmark built on 3D Gaussian Splatting indoor environments, designed to provide a unified platform for both policy evaluation and data generation under realistic closed-loop interaction. Beyond photorealistic egocentric RGB observations, GN-Bench supports metric-consistent BEV rendering, enabling a unified benchmark for both map-free and map-conditioned navigation paradigms. The simulator integrates rendering, motion execution, collision handling, and geodesic planning into a single interactive loop, allowing agents to be evaluated under physically constrained online execution rather than static offline replay. Importantly, GN-Bench also provides oracle planning and online relabeling capabilities, which make it possible to construct DAgger-style corrective supervision directly from policy rollouts. This allows the benchmark to serve not only as an evaluation substrate, but also as a practical training environment for studying recovery behaviors, covariate-shift mitigation, and closed-loop policy improvement in embodied navigation.

\begin{figure}[t]
    \centering
    \includegraphics[width=\linewidth]{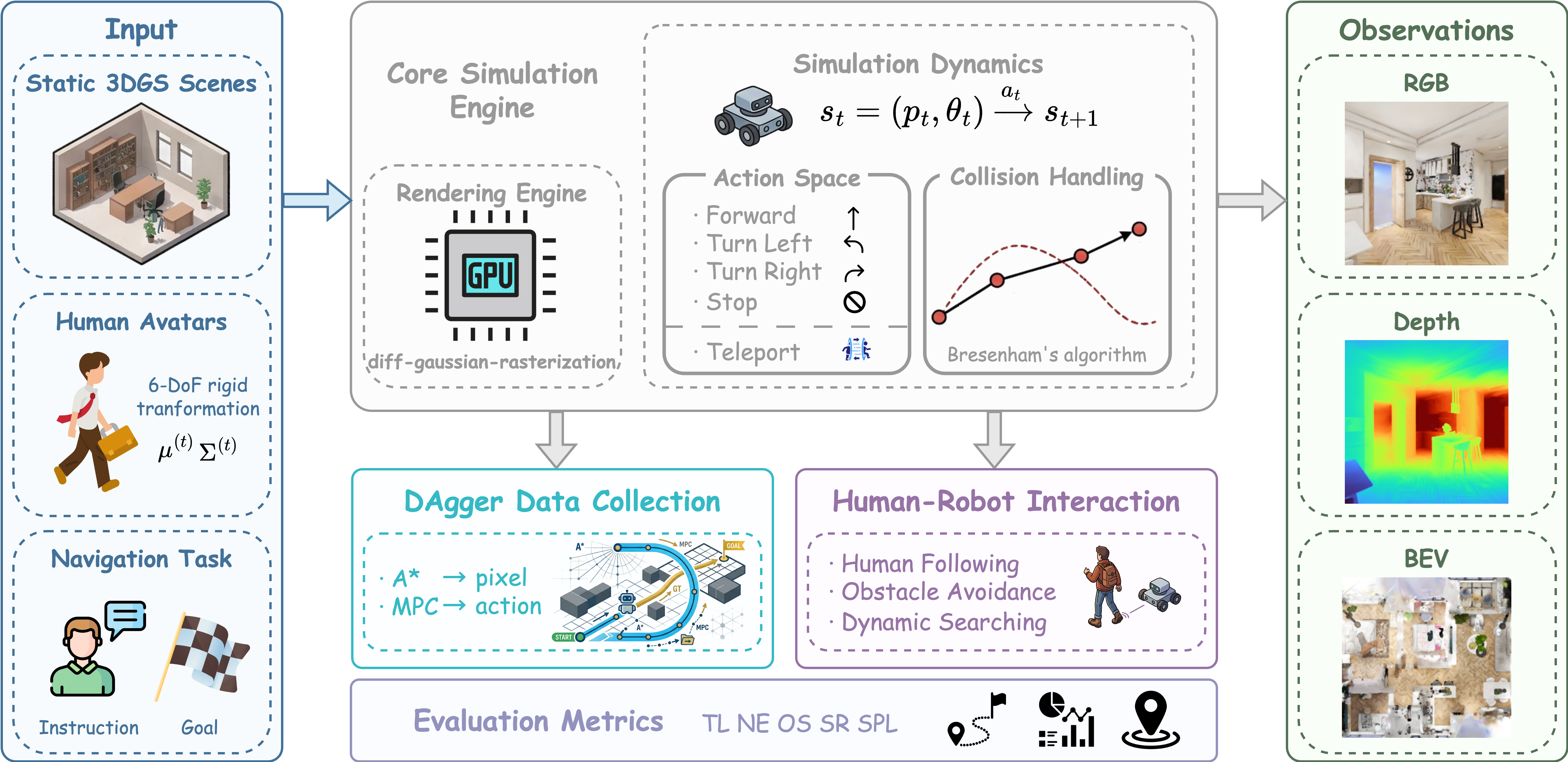}
    \caption{Overview of GN-Bench and its high-fidelity interactive simulation capabilities.}
    \label{fig:gnbench}
\end{figure}

\subsection{Core Rendering Engine}
GN-Bench leverages the highly optimized \texttt{diff-gaussian-rasterization} framework~\cite{kerbl3Dgaussians} as its core rendering engine. Designed to meet the high-throughput requirements of embodied AI research, this backend employs an efficient 3DGS pipeline to concurrently generate high-resolution RGB images and spatially aligned depth maps. Furthermore, this streamlined architecture facilitates cross-scene generalization, allowing standardized agent and sensor configurations to be seamlessly deployed across diverse environments through the direct instantiation of pre-trained Gaussian assets.


\subsection{Simulation Dynamics and Interactivity}
GN-Bench inherits and expands upon the robust simulation mechanics established by industry-standard platforms such as Habitat~\cite{savva2019habitat} and NavGSim~\cite{liu2026navgsimhighfidelitygaussiansplatting}. The platform is equipped with a comprehensive suite of embodied AI features:

\subsubsection{Observation Space} 
To support diverse embodied navigation paradigms, GN-Bench provides a comprehensive, multi-modal observation space, summarized in Table~\ref{tab:observation_space}. The simulator concurrently renders high-resolution egocentric RGB-D streams via perspective cameras and high-fidelity Bird's Eye View (BEV) projections. The BEV representations are synthesized by configuring a virtual downward-facing camera with an orthogonal projection matrix within the 3DGS rendering pipeline. This orthographic rasterization explicitly preserves metric scale by bypassing perspective foreshortening, thereby supplying a geometrically consistent spatial prior essential for global map-based planning.

\begin{table}[ht]
    \centering
    \caption{\textbf{The multi-modal observation space generated by GN-Bench.} The simulator supports the simultaneous rendering of egocentric and top-down perspectives to facilitate both map-free and map-based navigation paradigms. Image resolution ($H \times W$) and Field of View (FoV) are fully configurable hyperparameters.}
    \label{tab:observation_space}
    \resizebox{\columnwidth}{!}{
    \begin{tabular}{@{}llccl@{}}
        \toprule
        \textbf{Modality} & \textbf{Camera Type} & \textbf{Tensor Shape} & \textbf{FoV / Scale} & \textbf{Primary Downstream Utility} \\ \midrule
        Egocentric RGB & Perspective & $H \times W \times 3$ & Configurable & Visual semantic reasoning for VLMs \\
        Egocentric Depth & Perspective & $H \times W \times 1$ & Configurable & Local geometry \& reactive obstacle avoidance \\
        Top-down BEV & Orthographic & $H_{bev} \times W_{bev} \times C$ & Metric-scale & Global spatial prior \& topological planning \\ \bottomrule
    \end{tabular}
    }
\end{table}

\subsubsection{Action Space and Kinematics} 
To accommodate varying levels of navigational autonomy, the simulator defines the agent's spatial state at time step $t$ as $s_t = (p_t, \theta_t)$, where $p_t \in \mathbb{R}^2$ is the 2D coordinate and $\theta_t \in [-\pi, \pi)$ is the heading angle. The platform supports a hybrid action space $\mathcal{A} = \mathcal{A}_{local} \cup \mathcal{A}_{global}$. For fine-grained exploration, agents execute VLN-CE compliant discrete actions $a \in \mathcal{A}_{local}$ with fixed translation step $\Delta d$ and rotation angle $\Delta \theta$. The kinematic update function is formalized as:
\begin{equation}
    \label{eq:kinematics}
    s_{t+1} = 
    \begin{cases} 
        \big(p_t + \Delta d \cdot [\cos\theta_t, \sin\theta_t]^\top, \; \theta_t\big), & \text{if } a = \text{MoveForward} \\
        \big(p_t, \; \theta_t \pm \Delta \theta\big), & \text{if } a = \text{TurnLeft/Right} 
    \end{cases}
\end{equation}
Alternatively, to evaluate accelerated high-level waypoint planning, the system accepts global actions $a \in \mathcal{A}_{global}$. This enables direct spatial teleportation to explicitly defined global coordinates $p_{target}$, yielding an instantaneous state update $s_{t+1} = (p_{target}, \theta_{target})$. This dual formulation seamlessly bridges low-level continuous control with topological routing strategies.

\subsubsection{Collision Handling} 
To enforce rigorous physical constraints, GN-Bench implements a deterministic collision detection module based on the intrinsic grayscale occupancy maps from the GN-Matrix dataset. Let $\mathcal{M}$ denote the binary occupancy grid where obstacles are defined as $1$. To mathematically account for the agent's physical footprint of radius $r$, we apply a morphological dilation using a circular structuring element $\mathcal{B}_r$. This yields the expanded obstacle map $\mathcal{M}_{dil} = \mathcal{M} \oplus \mathcal{B}_r$, which effectively transforms the collision boundaries relative to the agent's center of mass. For any proposed kinematic translation from a source coordinate $p_{src}$ to a target $p_{dst}$, the trajectory is discretized into a finite sequence of grid pixels via Bresenham's line algorithm~\cite{bresenham}. This sequence, denoted as $\mathcal{P} = (p_0, p_1, \dots, p_N)$ with $p_0 = p_{src}$ and $p_N = p_{dst}$, is sequentially evaluated against $\mathcal{M}_{dil}$. If the path intersects an occupied cell, the movement vector is strictly truncated. The agent's spatial state is then updated to the final collision-free coordinate $p_{final} = p_k$ along the intended trajectory, formalized as:
\begin{equation}
    \label{eq:collision}
    k = \max \{ i \in [0, N] \mid \mathcal{M}_{dil}(p_j) = 0, \; \forall j \le i \}
\end{equation}
As expressed in Eq.~\ref{eq:collision}, this discrete formulation ensures penetration-free continuous roaming while maintaining high computational efficiency during large-scale simulation.

\subsubsection{Global Path Planning} 
GN-Bench incorporates an oracle path planner to facilitate goal-oriented navigation and standardized benchmarking. The system applies the A* heuristic search algorithm~\cite{4082128} directly to the dilated occupancy grids. Because these grids explicitly account for the agent's physical footprint, the planner reliably computes optimal, collision-free trajectories between any traversable states. This deterministic formulation yields the ground-truth geodesic distances required to evaluate core metrics such as Success weighted by Path Length (SPL)~\cite{anderson2018evaluation}. Concurrently, it acts as an expert demonstrator to synthesize the oracle annotations required for our DAgger data collection pipeline (as described in Sec.~\ref{sec:dagger}), thereby providing direct supervision for the training of map-based navigation policies.

\subsection{Dynamic 3DGS Avatars and Human-Robot Interaction}
A critical limitation of conventional VLN simulators is their reliance on strictly static environments. To address this, GN-Bench seamlessly integrates dynamic 3DGS human avatars into the foundational GN-Matrix scenes, thereby synthesizing highly realistic, non-stationary navigation scenarios. 

From a rendering perspective, let $\mathcal{G}_{static}$ denote the set of 3D Gaussians representing the static indoor environment, and $\mathcal{G}_{avatar}$ denote the Gaussians of the human avatar. At each simulation time step $t$, the avatar is driven by a spatiotemporal trajectory specifying its 6-DoF pose, defined by a rotation matrix $R_t$ and a translation vector $T_t$. To dynamically render the human within the scene, we apply a rigid transformation to the mean $\mu_i$ and covariance $\Sigma_i$ of each Gaussian $i \in \mathcal{G}_{avatar}$:
\begin{equation}
    \label{eq:dynamic_gs}
    \mu_i^{(t)} = R_t \mu_i + T_t, \quad \Sigma_i^{(t)} = R_t \Sigma_i R_t^\top
\end{equation}
The updated avatar Gaussians $\mathcal{G}_{avatar}^{(t)}$ are then concatenated with $\mathcal{G}_{static}$ before the unified rasterization pass. By introducing these spatiotemporally varying human subjects, our platform establishes a novel paradigm for evaluating dynamic Human-Robot Interaction (HRI). This formulation inherently supports the rigorous benchmarking of complex, time-critical embodied tasks, specifically active target following and reactive dynamic obstacle avoidance. Consequently, GN-Bench extends standard VLN evaluation protocols to better reflect the unpredictability of real-world robotic deployment in populated spaces.

\section{GN-BAE}

Break and Establish (BAE) is a unified embodied navigation foundation model for instruction following, human following, and goal-directed navigation, following the general formulation of vision-and-language navigation and continuous embodied navigation~\cite{anderson2018vision,krantz2020beyond}. It supports FPV, BEV, and hybrid visual inputs, while jointly modeling both discrete action sequences and continuous trajectories within a single framework. At its core, BAE adopts a trajectory tokenization scheme that unifies planning representation and improves training efficiency across diverse navigation settings. Combined with high-fidelity rendered data~\cite{kerbl3Dgaussians} and supervised fine-tuning, this design effectively unlocks the navigation capability of the underlying vision-language model. A key component is our closed-loop post-training pipeline: after supervised learning, DAgger~\cite{ross2011reduction} exposes the model to rollout-induced states and recovery behaviors beyond idealized offline demonstrations, breaking the narrow expert-centric distribution inherited from supervision and providing a better starting point for downstream exploration. Building on this initialization, DAPO~\cite{yu2025dapo} further refines the policy with execution-oriented rewards, enabling more effective exploration, stronger long-horizon decision-making, and more reliable action generation during rollout. Lightweight Action Experts further bridge high-level planning and executable control for practical deployment~\cite{cai2025navdp}.

The model architecture is composed of five components. During Vision-and-Language Navigation (VLN) training~\cite{anderson2018vision,krantz2020beyond}, coordinate tokenization significantly enhances efficiency, simplifies the learning process, and unifies task execution. By leveraging GS high-fidelity rendering~\cite{kerbl3Dgaussians} with Supervised Fine-Tuning (SFT), we can activate the latent pre-trained knowledge of Qwen3-VL~\cite{Qwen3-VL}, preserving as much of the pre-training knowledge as possible. In the reinforcement learning process, increasing entropy helps expand the exploration range of the policy, thereby improving the diversity and scope of learning. Furthermore, on the RL-trained Vision-and-Language Model (VLM), we employ Action Expert~\cite{cai2025navdp} to adapt to different ontologies, enabling cross-ontology capability transfer while significantly improving edge inference efficiency. The combination of these techniques results in a more efficient training process, providing enhanced adaptability and superior inference performance in real-world applications.

The tasks include a variety of capabilities, such as instruction following, human following, and goal navigation. The system is designed to handle different types of visual input, including Bird’s Eye View (BEV), First-Person View (FPV), or a combination of both BEV and FPV. Additionally, the output trajectories can be represented in two formats: discrete actions, which correspond to specific movement commands, or continuous trajectory points, which provide a denser representation of the planned path. This flexibility in both input modalities and output formats enables the model to support a broad range of navigation scenarios under a unified framework.

For FPV-based navigation, we introduce a simple memory mechanism to compensate for the lack of global spatial context. Specifically, the current frame observation is treated as short-term memory, while the observations associated with the previous sixteen executed discrete actions are used as long-term memory. These sixteen historical observations are concatenated into a single image and fed into the model together with the current frame. As a result, the FPV input contains not only the current egocentric view, but also a compact visual history aligned with the recent action sequence, which improves temporal reasoning and action prediction.

\begin{figure}[t]
\centering
\includegraphics[width=\linewidth]{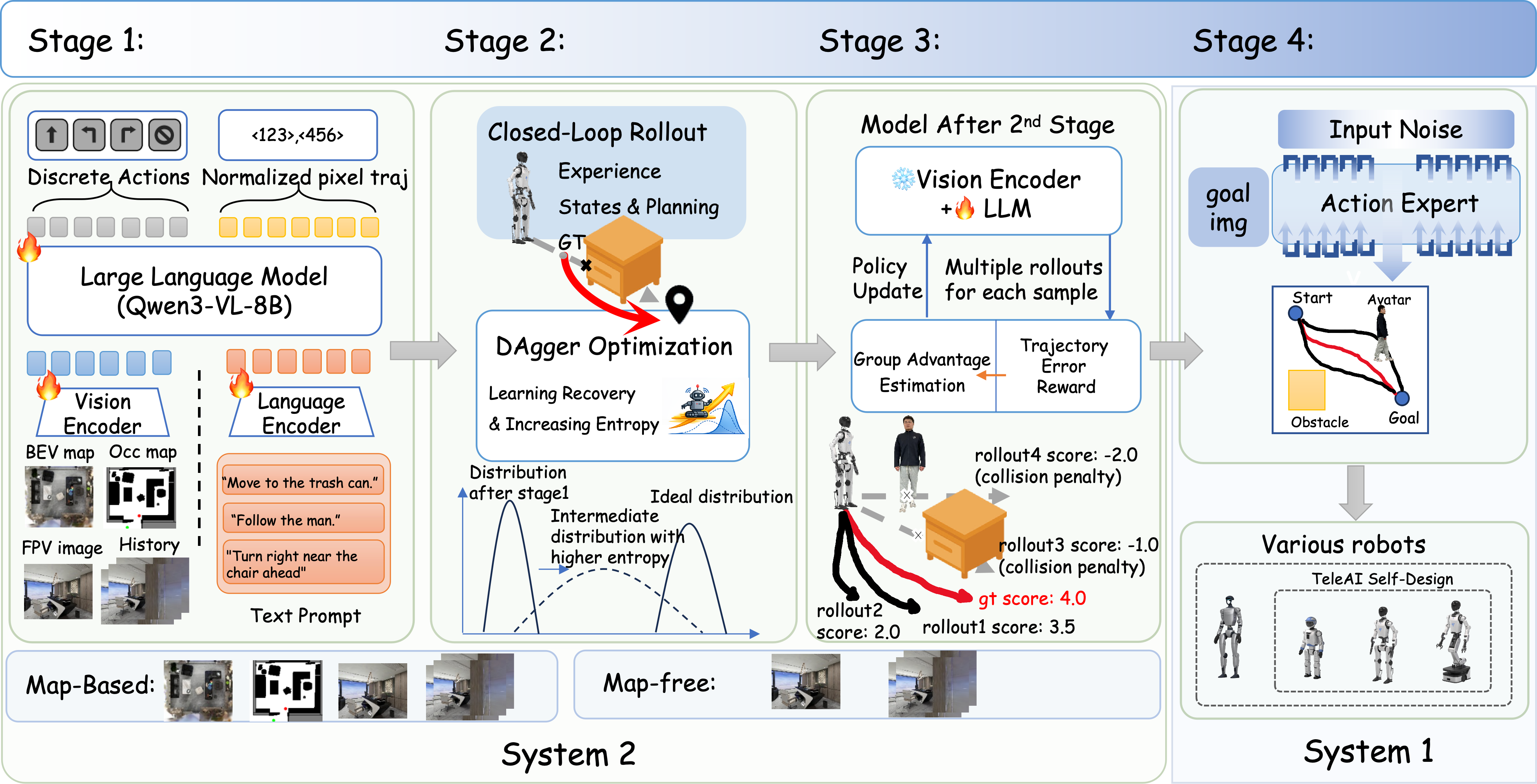}
\caption{Architecture of BAE and its Four-Stage Training Strategy of DAgger + RL.}
\label{fig:model_architecture}
\end{figure}

\subsection{Trajectory tokenization}

We formulate trajectory prediction as discrete sequence generation instead of direct continuous regression.
For autoregressive VLMs, discrete token prediction is generally easier to optimize, more stable across heterogeneous tasks, and more convenient for unifying different output ontologies.
Inspired by sequence-based visual prediction and location-token modeling~\cite{chen2022pix2seq,chen2022unified,peng2024kosmos2,jiang2025rexomni}, we adopt a tokenized trajectory representation for both discrete actions and continuous waypoints.
In particular, following the idea of relative coordinates with special tokens, we map continuous spatial outputs into a bounded discrete token space, which reduces learning difficulty and improves token efficiency.

For trajectories represented by discrete actions, we define a lightweight action vocabulary consisting of \texttt{<action>}, \texttt{</action>}, \texttt{<FWD>}, \texttt{<LEFT>}, \texttt{<RIGHT>}, and \texttt{<STOP>}.
Here, \texttt{<action>} and \texttt{</action>} indicate the action span, while the four atomic action tokens follow the VLN-CE action semantics~\cite{krantz2020beyond}: \texttt{<FWD>} moves the agent forward by 0.25 meters, \texttt{<LEFT>} rotates the agent left by 15 degrees, \texttt{<RIGHT>} rotates the agent right by 15 degrees, and \texttt{<STOP>} terminates navigation.
This design converts low-level control decisions into a compact symbolic sequence and allows the model to learn action policies through standard next-token prediction.

For trajectories represented by continuous points, we discretize normalized BEV coordinates into 1,000 bins and assign one special token to each bin, i.e., \texttt{<0>} to \texttt{<999>}.
Each waypoint is first projected onto the BEV plane and then normalized with respect to the current BEV canvas size, so that all coordinates are represented in a shared token space.
This is particularly useful in our setting because BEV images may have different resolutions across datasets and tasks.
Without normalization, the same physical location could correspond to different pixel values, which increases the learning burden.
By contrast, relative coordinates with special tokens provide a unified and resolution-agnostic representation, while also requiring fewer output tokens than coordinate strings or digit-level serialization.

With this design, BAE supports both action-sequence prediction and waypoint-sequence prediction under the same autoregressive modeling framework, providing a common representation layer across different navigation tasks and deployment ontologies.

\subsection{VLM SFT}

During the SFT stage, we train BAE with standard supervised fine-tuning on trajectory-token prediction.
Given a multimodal input $x$ and a target trajectory token sequence $y^*$, the training objective is the autoregressive cross-entropy loss:
\begin{equation}
\mathcal{L}_{\mathrm{SFT}}(\theta)
=
-\mathbb{E}_{(x,y^*)\sim\mathcal{D}}
\left[
\frac{1}{L}
\sum_{t=1}^{L}
\log \pi_{\theta}\left(y_t^* \mid y_{<t}^*, x\right)
\right].
\end{equation}

The supervision targets are constructed from both discrete actions and pixel-space trajectories.
Specifically, we use the next six discrete actions as the action label sequence, and a future 3-meter trajectory in the BEV space as the continuous trajectory label.
The 3-meter trajectory is uniformly discretized into 12 pixel points, which are then converted into trajectory tokens using the tokenization scheme described above.
For pure FPV input, we do not apply the pixel-trajectory output branch, and the training target only contains the discrete action sequence.

We jointly finetune the entire model, including the vision encoder, the multimodal projector, and the language model.
To stabilize training, we use a smaller learning rate for the vision tower ($2\times10^{-6}$), while setting the learning rates of the multimodal projector and the LLM to $1\times10^{-5}$.
This imitation learning stage establishes the basic planning capability of the VLM for downstream navigation tasks.

\subsection{DAgger-based Closed-loop Adaptation}

A policy obtained from pure supervised fine-tuning is often overly concentrated around idealized expert trajectories, which may be suboptimal for subsequent reinforcement learning.
In our setting, this issue is further exacerbated because the SFT targets are derived from ideal A* trajectories~\cite{4082128}.
While such supervision is effective for learning canonical navigation behaviors, it provides limited coverage of recovery actions under closed-loop execution, where the agent may deviate from the intended path and must correct errors induced by its own previous decisions.
As a result, a model initialized solely from SFT may have limited support over the state-action distribution encountered at the beginning of GRPO training~\cite{shao2024deepseekmathpushinglimitsmathematical}.

To mitigate this mismatch, we introduce an additional DAgger stage before GRPO~\cite{ross2011reduction,shao2024deepseekmathpushinglimitsmathematical}:
\begin{equation}
\mathcal{L}_{\mathrm{DAgger}}(\theta)
=
-\mathbb{E}_{(x,y^*)\sim\mathcal{D}_{\mathrm{DAgger}}}
\left[
\frac{1}{L}
\sum_{t=1}^{L}
\log \pi_{\theta}\left(y_t^* \mid y_{<t}^*, x\right)
\right].
\end{equation}

In practice, we collect 284k closed-loop DAgger samples and optimize the model for 5 epochs with a learning rate of $1\times10^{-6}$.
During this stage, we continue updating the full VLM, including the vision encoder, rather than freezing the visual backbone.
Importantly, this stage is not intended to drive the training loss to a very low level.
Instead, its purpose is to adapt the post-SFT policy toward the state distribution induced by real closed-loop rollouts, where trajectory deviations and corrective behaviors naturally arise.

This design serves two closely related goals.
First, DAgger improves the model's ability to recover from realistic closed-loop perturbations, which is beneficial to navigation performance in its own right.
More importantly, it provides a better initialization for GRPO.
From a distributional perspective, DAgger moves the policy away from the overly concentrated SFT solution toward an intermediate distribution that better balances ideal expert behavior and the empirical state distribution encountered during closed-loop execution.
Such an initialization increases the likelihood of sampling reasonable trajectories in the early stage of DAPO, and thereby facilitates more effective exploration and optimization in subsequent reinforcement learning.

\subsection{DAPO}

After the DAgger stage, we further optimize BAE with DAPO under closed-loop interaction~\cite{yu2025dapo}.
Although DAgger already improves the policy's robustness to the state distribution induced by its own actions, it remains fundamentally an imitation-style training stage~\cite{ross2011reduction}.
To further improve long-horizon decision quality and directly optimize execution-oriented objectives, we introduce a reinforcement learning stage built upon the DAgger-initialized model.

Before DAPO training, we first use the post-DAgger model to interact with the environment in a closed-loop manner, following a data collection procedure similar to that used in the DAgger stage.
In this way, we collect an additional 22k closed-loop samples.
These samples are not used as supervised labels; instead, they serve as the reference basis for reward computation during DAPO training.
Compared with directly starting reinforcement learning from the pure SFT model, this design produces an initial rollout distribution that is substantially closer to realistic execution trajectories, thereby improving reward reliability and stabilizing the early stage of policy optimization.

For policy optimization, we adopt the GRPO-style advantage estimator~\cite{shao2024deepseekmathpushinglimitsmathematical} within the DAPO framework~\cite{yu2025dapo}.
Given an input $x$, the current policy $\pi_\theta$ samples a group of $G$ candidate responses
$\{y_1, y_2, \dots, y_G\}$,
where each response corresponds to a complete predicted trajectory.
For each sampled response $y_i$, we compute a scalar reward $r_i$ based on the execution-oriented reward defined below.
The reward is then normalized within the group to obtain a relative advantage,
\begin{equation}
A_i
=
\frac{r_i - \mathrm{mean}(r_1,\dots,r_G)}
{\mathrm{std}(r_1,\dots,r_G)}.
\end{equation}
In this way, policy optimization is driven by relative trajectory quality within each sampled group, so that trajectories with better rollout outcomes receive larger positive updates, while worse trajectories are suppressed.

Following the DAPO/GRPO formulation~\cite{shao2024deepseekmathpushinglimitsmathematical,yu2025dapo}, we optimize the policy with a clipped policy-gradient objective~\cite{schulman2017proximal}:
\begin{equation}
\mathcal{J}_{\mathrm{DAPO}}(\theta)
=
\frac{1}{G}
\sum_{i=1}^{G}
\frac{1}{|y_i|}
\sum_{t=1}^{|y_i|}
\min\left(
\rho_{i,t} A_i,\;
\mathrm{clip}(\rho_{i,t}, 1-\epsilon_{\mathrm{low}}, 1+\epsilon_{\mathrm{high}}) A_i
\right),
\end{equation}
where
\begin{equation}
\rho_{i,t}
=
\frac{\pi_\theta(y_{i,t}\mid x, y_{i,<t})}
{\pi_{\theta_{\mathrm{old}}}(y_{i,t}\mid x, y_{i,<t})}
\end{equation}
is the importance ratio between the updated and behavior policy at token step $t$.
This objective preserves the core idea of policy-gradient optimization while preventing excessively large updates through ratio clipping.

The necessity of DAPO in our setting can be understood from two aspects.
First, under standard SFT, trajectory generation is optimized only through teacher-forced next-token prediction.
Such supervision is effective for learning the required format and canonical expert behavior, but it does not distinguish whether a sampled trajectory is globally better at the rollout level.
By contrast, DAPO allows gradient updates to be weighted by trajectory-level quality, rather than treating all target tokens equally under next-token prediction.
Second, this design is especially important for our normalized pixel branch.
Although normalized coordinate tokens provide an efficient representation for waypoint prediction, they also introduce a mismatch between discrete token supervision and continuous geometric quality.
Two predictions that differ by only one coordinate token may correspond to a negligible spatial deviation, while predictions with similar token-level losses may lead to substantially different rollout quality.
Therefore, for the normalized pixel branch, reinforcement learning is necessary to bridge this discrete-to-continuous gap by directly optimizing geometry-aware trajectory similarity, rather than relying solely on token-level supervision.

In our implementation, each prompt samples $8$ candidate responses during rollout.
We disable KL regularization in both the actor loss and the reward term, and optimize the model with a learning rate of $1\times10^{-6}$.
This choice is made empirically: we observe that enabling KL loss can easily destabilize optimization in tasks with strong output-format constraints, and in severe cases may even cause the policy to collapse with respect to the required action-generation format.
In our setting, such regularization is unnecessary, since format validity is already explicitly enforced by the reward design, which assigns large penalties to invalid predictions and therefore serves as an effective safeguard for structured generation.
During this stage, we freeze the vision tower and update only the remaining trainable components of the VLM.
After SFT and DAgger, the visual backbone is already sufficiently adapted to the rendered navigation observations, while the primary role of reinforcement learning is to further refine action generation behavior rather than relearn visual perception.

The reward function is designed as an execution-oriented structured score.
Given a predicted action sequence $\hat{a}$ and its reference action sequence $a^*$, we define the overall reward as
\begin{equation}
R(\hat{a}, a^*)
=
R_{\mathrm{traj}}
+
\lambda_{\mathrm{first}} R_{\mathrm{first}}
-
\lambda_{\mathrm{col}} R_{\mathrm{col}}
-
\lambda_{\mathrm{len}} R_{\mathrm{len}}
-
\lambda_{\mathrm{fmt}} R_{\mathrm{fmt}},
\end{equation}
where $R_{\mathrm{traj}}$ measures rollout-level consistency between the predicted and reference trajectories under shared initial conditions, $R_{\mathrm{first}}$ rewards correct prediction of the first action, $R_{\mathrm{col}}$ penalizes collision during rollout, $R_{\mathrm{len}}$ penalizes length mismatch between the predicted and reference action sequences, and $R_{\mathrm{fmt}}$ denotes the format penalty for invalid structured outputs.
The trajectory term jointly considers both positional consistency and heading consistency under closed-loop rollout.
Overall, this reward encourages the model to generate responses that are not only structurally valid, but also dynamically consistent with the reference behavior and executable in the environment.

In practice, the DAPO stage is trained with a batch size of $64$, while $8$ responses are sampled for each prompt during rollout.
We optimize the model for $2$ epochs.

Moreover, our ablation study further shows that directly applying DAPO without the DAgger stage does not lead to meaningful performance improvement.
A plausible reason is that, after Stage 1, the policy entropy is already very low, which severely limits exploration at the beginning of reinforcement learning and consequently prevents DAPO from yielding substantial capability gains.

\subsection{Action Expert}

For waypoint-based outputs, we do not directly execute the predicted pixel trajectory.
Instead, we first transform the predicted BEV trajectory into the agent-centered first-person world coordinate frame, and then feed it into a dedicated NavDP Action Expert~\cite{cai2025navdp} to generate executable low-level actions.
We train the Action Expert from scratch following the training pipeline and loss design of the InternNav repository.
In practice, we maintain two separate experts for different task families, namely a follow expert and a planning expert.
The Action Expert is introduced not only to produce smoother and more stable action execution, but also to improve safety in cluttered environments by providing stronger local obstacle avoidance capability.
In this way, the VLM is responsible for high-level trajectory planning, while the Action Expert serves as a trajectory-to-action adapter that improves execution robustness and safety.

To better align the Action Expert with the VLM input modality, we remove the depth input from the original NavDP design~\cite{cai2025navdp} and use only visual observations and goal-related conditions.
NavDP is trained with a DDPM-style noise prediction objective~\cite{ho2020denoising}, following diffusion-policy-based visuomotor learning~\cite{chi2023diffusionpolicy,cai2025navdp}, where Gaussian noise is added to the target action trajectory and the model learns to predict the injected noise:
\begin{equation}
\mathbf{a}_{\tau}
=
\sqrt{\bar{\alpha}_{\tau}}\,\mathbf{a}
+
\sqrt{1-\bar{\alpha}_{\tau}}\,\boldsymbol{\epsilon},
\qquad
\boldsymbol{\epsilon}\sim\mathcal{N}(\mathbf{0},\mathbf{I}),
\end{equation}
\begin{equation}
\mathcal{L}_{\mathrm{act}}
=
\mathbb{E}
\left[
\left\|
\hat{\boldsymbol{\epsilon}}_{\theta}(\mathbf{a}_{\tau}, \tau, \cdot)
-
\boldsymbol{\epsilon}
\right\|_2^2
\right].
\end{equation}
In addition to the diffusion action loss, the training objective also includes a critic regression term for obstacle-aware trajectory quality estimation and an auxiliary cross-modal goal alignment term~\cite{cai2025navdp}.
The overall loss is
\begin{equation}
\mathcal{L}
=
0.8\,\mathcal{L}_{\mathrm{act}}
+
0.2\,\mathcal{L}_{\mathrm{critic}}
+
0.5\,\mathcal{L}_{\mathrm{aux}}.
\end{equation}

\section{Experiments}

\subsection{Experiment Setup}

\textbf{Simulated environments.}
We conduct evaluations across three distinct simulation platforms. \textbf{GN-Bench} serves as our primary high-fidelity, 3DGS-based continuous navigation environment. \textbf{VLN-CE} provides a standard benchmark built upon the Habitat simulator~\cite{savva2019habitat} and Matterport3D scenes, focusing on natural language instruction following under continuous control. Finally, \textbf{VLN-PE} offers a physics-aware framework that accurately models the real-world dynamics and low-level control errors of a Unitree H1 humanoid robot, bridging the sim-to-real gap.

\textbf{Real-world environments.} To evaluate the model’s performance in real-world environments, we conducted three experiments across multiple robotic platforms, including a self-developed wheeled–arm robot TeleBotW, a self-developed small humanoid robot TeleBotM, and the Unitree Robotics G1 humanoid robot. The system was deployed on an NVIDIA Orin edge device, while inference was executed on an NVIDIA H100 GPU, achieving a real-time inference frequency of up to 5 Hz.

\textbf{Metrics.}
To comprehensively evaluate navigation performance in continuous and physically realistic environments, we adopt standard metrics for both task completion and kinematic stability. Specifically, we measure Navigation Error (NE), defined as the terminal Euclidean distance to the goal; Success Rate (SR), the percentage of episodes where the agent halts within a 3-meter success radius; and Oracle Success (OS), indicating whether any point along the agent's trajectory breaches the success threshold. To quantify path efficiency, we report Trajectory Length (TL) as the total physical distance traversed, alongside Success weighted by Path Length (SPL), which heavily penalizes detours. 

\textbf{Baselines.}
We compare GN-BAE against a diverse spectrum of state-of-the-art visual navigation methods. These encompass traditional multi-sensor paradigms reliant on panoramic RGB, depth, and odometry (e.g., CMA, ETPNav), VLM-free single-view RGB-D approaches, and recent Video-LLM architectures restricted solely to egocentric RGB inputs (e.g., NaVid, InternNav).

\begin{table}[b]
    \caption{\textbf{Evaluation Metrics on GN-Bench.} In the observation space, \textbf{Depth} denotes depth maps, \textbf{BEV} refers to Bird's Eye View projections, and \textbf{FPV} represents first-person RGB images. \textbf{Bold} represents the best results. \uline{Underline} indicates the second best results. $^{\dagger}$ Indicates methods fine-tuned via SFT on the NavGMatrix dataset.}
    \label{tab:GN-Bench}
    \centering
    \setlength{\tabcolsep}{4.35pt}
    \begin{NiceTabular}{lccccccccccccc}
        \CodeBefore
            \columncolor[HTML]{EFEFEF}{2-4}
            \rowcolor[HTML]{F0F8FF}{10-11}
        \Body
        \toprule
        \multirow{2}{*}{\textbf{Method}} & \multicolumn{3}{c}{\textbf{Observation}} & \multicolumn{5}{c}{\textbf{Metrics on Seen}} & \multicolumn{5}{c}{\textbf{Metrics on Unseen}} \\ 
        \cmidrule(lr){2-4} \cmidrule(lr){5-9} \cmidrule(lr){10-14}
         & \textbf{Depth} & \textbf{BEV} & \textbf{FPV} & \textbf{TL} & \textbf{NE} $\downarrow$ & \textbf{OS} $\uparrow$ & \textbf{SR} $\uparrow$ & \textbf{SPL} $\uparrow$ & \textbf{TL} & \textbf{NE} $\downarrow$ & \textbf{OS} $\uparrow$ & \textbf{SR} $\uparrow$ & \textbf{SPL} $\uparrow$ \\ 
        \midrule
        CMA & \checkmark &  & \checkmark & 2.9 & 8.3 & 15.7 & 12.5 & 11.9 & 3.0 & 8.1 & 19.6 & 15.5 & 14.9 \\
        NaVid & &  & \checkmark & 3.4 & 7.9 & 20.1 & 14.6 & 12.8 & 3.5 & 7.7 & 20.3 & 14.5 & 12.8 \\
        UniNaVid &  &  & \checkmark & 4.6 & 7.9 & 22.2 & 15.0 & 12.5 & 5.2 & 7.8 & 20.7 & 12.8 & 10.3 \\
        InternNav(S2) &  &  & \checkmark & 3.6 & 7.4 & 23.1 & 18.8 & 17.5 & 3.7 & 7.2 & 26.7 & 22.1 & 20.3 \\
        NaVid$^\dagger$ &  &  & \checkmark & 2.7 & 7.4 & 19.4 & 18.8 & 18.8 & 2.7 & 7.1 & 23.8 & 23.1 & 23.0 \\
        UniNaVid$^\dagger$ &  &  & \checkmark & 3.7 & 7.2 & 24.1 & 22.5 & 21.9 & 5.8 & 7.5 & 23.1 & 20.8 & 20.2 \\
        InternNav(S2)$^\dagger$ &  &  & \checkmark & 2.9 & 7.1 & 22.5 & 22.4 & 22.4 & 2.9 & 6.9 & 24.9 & 24.0 & 23.7 \\
        \midrule
        \textbf{GN-BAE} &  &  & \checkmark & 5.2 & \underline{4.9} & \underline{48.9} & \underline{46.4} & \underline{44.7} & 5.0 & \textbf{5.6} & \textbf{43.6} & \textbf{38.9} & \underline{37.3} \\ 
        \textbf{GN-BAE} &  & \checkmark & \checkmark & 5.2 & \textbf{4.3} & \textbf{59.3} & \textbf{58.6} & \textbf{58.6} & 4.0 & \underline{5.8} & \underline{40.2} & \underline{38.5} & \textbf{38.2} \\ 
        \bottomrule
    \end{NiceTabular}
\end{table}

\subsection{Simulated Environment}

\subsubsection{Evaluation on GN-Bench}
We first evaluate GN-BAE on the GN-Bench benchmark (Table~\ref{tab:GN-Bench}). While traditional methods rely on auxiliary depth, recent VLM-based baselines and our approach operate strictly on monocular RGB (FPV) or Bird's Eye View (BEV). Despite this minimal sensory constraint, GN-BAE establishes a new state-of-the-art by a massive margin. Notably, on the Unseen split, our FPV-only variant achieves a 38.9\% SR, outperforming the strongest fine-tuned baseline InternNav-S2$^\dagger$ by nearly 15 absolute points, demonstrating exceptional spatial reasoning in a low-level action space.

Comparing our two variants reveals an insightful trade-off between modality and generalization. On the Seen split, the BEV-augmented model dominates, indicating that global geometric priors substantially facilitate navigation in familiar environments. However, on the Unseen split, the FPV-only variant yields better NE, OS, and SR, while the BEV variant only retains a marginal edge in SPL. This suggests that while BEV improves path efficiency, imperfect map cues in novel 3DGS environments can introduce representation bias, occasionally misleading long-horizon decisions. Conversely, the FPV-only policy relies on robust egocentric visual-language grounding, proving to be more resilient to scene-level distribution shifts for goal reaching.

\begin{table}[t]
    \caption{\textbf{Comparison with state-of-the-art methods on VLN-CE R2R Val-Unseen split.} * indicates methods using the waypoint predictor from Hong et al. (2022).}
    \label{tab:VLN-CE}
    \centering
    \small
    \resizebox{1.0\textwidth}{!}{
        \begin{NiceTabular}{lcccccccc}
        \CodeBefore
            \columncolor[HTML]{EFEFEF}{2-5} 
            \rowcolor[HTML]{F0F8FF}{23} 
        \Body
        \toprule
        \multirow{2}{*}{\textbf{Method}} & \multicolumn{4}{c}{\textbf{Observation}} & \multicolumn{4}{c}{\textbf{R2R Val-Unseen}} \\ 
        \cmidrule(lr){2-5}
        \cmidrule(lr){6-9} 
         & \textbf{Pano.} & \textbf{Odo.} & \textbf{Depth} & \textbf{S.RGB} & \textbf{NE} $\downarrow$ & \textbf{OS} $\uparrow$ & \textbf{SR} $\uparrow$ & \textbf{SPL} $\uparrow$ \\ 
         \midrule
        HPN+DN* & \checkmark & \checkmark & \checkmark & & 6.31 & 40.0 & 36.0 & 34.0 \\
        CMA* & \checkmark & \checkmark & \checkmark & & 6.20 & 52.0 & 41.0 & 36.0 \\
        GridMM* & \checkmark & \checkmark & \checkmark & & 5.11 & 61.0 & 49.0 & 41.0 \\
        ETPNav* & \checkmark & \checkmark & \checkmark & & 4.71 & 65.0 & 57.0 & 49.0 \\
        ScaleVLN* & \checkmark & \checkmark & \checkmark & & 4.80 & - & 55.0 & 51.0 \\
        InstructNav & \checkmark & \checkmark & \checkmark & \checkmark & 6.89 & - & 31.0 & 24.0 \\
        R2R-CMTP & \checkmark & \checkmark & \checkmark & & 7.90 & 38.0 & 26.4 & 22.7 \\
        LAW & & \checkmark & \checkmark & \checkmark & 6.83 & 44.0 & 35.0 & 31.0 \\
        CM2 & & \checkmark & \checkmark & \checkmark & 7.02 & 41.5 & 34.3 & 27.6 \\
        WS-MGMap & & \checkmark & \checkmark & \checkmark & 6.28 & 47.6 & 38.9 & 34.3 \\
        ETPNav + FF & & \checkmark & \checkmark & \checkmark & 5.95 & 55.8 & 44.9 & 30.4 \\
        Seq2Seq & & & \checkmark & \checkmark & 7.77 & 37.0 & 25.0 & 22.0 \\
        CMA & & & \checkmark & \checkmark & 7.37 & 40.0 & 32.0 & 30.0 \\
        NaVid & & & & \checkmark & 5.47 & 49.1 & 37.4 & 35.9 \\
        MapNav & & & & \checkmark & 4.93 & 53.0 & 39.7 & 37.2 \\
        NaVILA & & & & \checkmark & 5.22 & 62.5 & 54.0 & 49.0 \\
        UniNaVid & & & & \checkmark & 5.58 & 53.3 & 47.0 & 42.7 \\
        StreamVLN & & & & \checkmark & 4.98 & 64.2 & 56.9 & 51.9 \\ 
        DualVLN & & & & \checkmark & 4.05 & 70.7 & 64.3 & 58.5 \\
        CorrectNav & & & & \checkmark & 4.24 & 67.5 & 65.1 & 62.3 \\
        \midrule
        \textbf{GN-BAE} & & & & \checkmark & \textbf{3.50} & \textbf{75.6} & \textbf{67.7} & \textbf{63.4} \\ 
        \bottomrule
        \end{NiceTabular}
    }
\end{table}

\subsubsection{Evaluation on VLN-CE}
We evaluate GN-BAE on the standard R2R-CE benchmark using the Validation Unseen split. As shown in Table~\ref{tab:VLN-CE}, GN-BAE achieves the best overall performance, reaching 3.50 NE, 75.6 OS, 67.7 SR, and 63.4 SPL using only single-view RGB input. Compared with the strongest prior RGB-only baseline, GN-BAE reduces NE by 17.5\%(from 4.24 to 3.50m) while improving SR and SPL by 2.6 and 1.1 points, respectively, indicating better goal localization and trajectory efficiency.

We attribute this improvement to the proposed offline reinforcement learning paradigm. Unlike conventional SFT-based navigation models, which mainly learn to imitate token-level action labels, GN-BAE further optimizes the policy distribution toward high-quality navigation trajectories. In particular, the DAgger-style stage exposes the model to states induced by its own policy, mitigating covariate shift caused by long-horizon closed-loop execution. The subsequent offline RL stage then encourages actions that are more consistent with successful and efficient navigation behavior, rather than merely matching the marginal action distribution of the training data. As a result, the learned policy better aligns with the ideal navigation distribution, leading to stronger generalization in unseen environments.

\subsection{Experiment in Real-world Environment}

\begin{figure}[t]
    \centering
    \includegraphics[width=\linewidth]{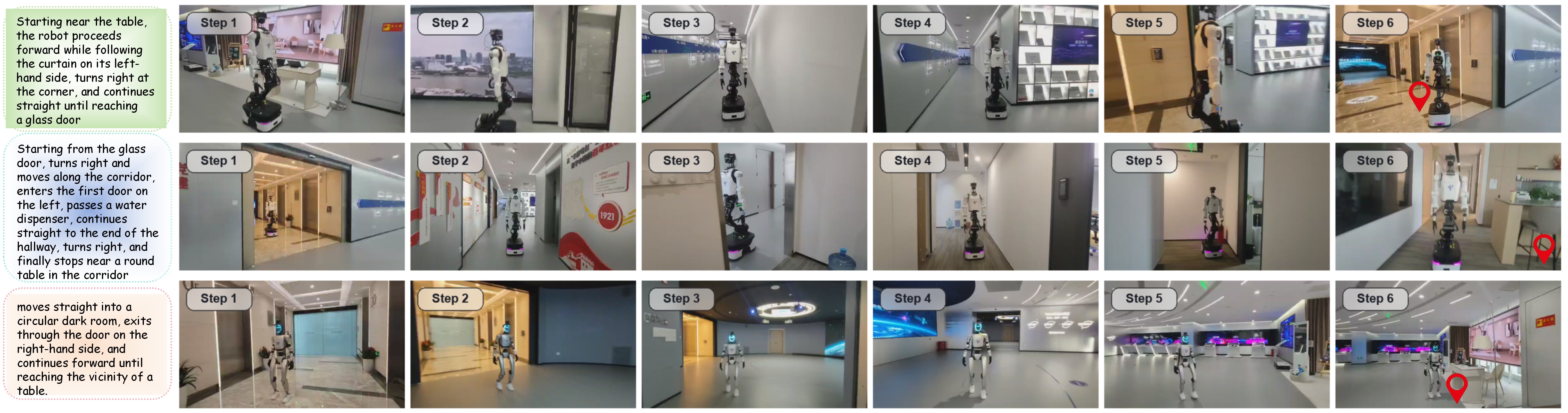}
    \caption{Experiment in Real-world Environment with self-developed wheeled–arm robot TeleBotW and Unitree Robotics G1 humanoid robot.}
    \label{fig:realworld-exp}
\end{figure}

Trained solely on high-fidelity Gaussian Splatting (GS) simulation data, our model demonstrates strong sim-to-real generalization capability. Notably, the model is able to accomplish real-world tasks reliably without any exposure to real robot data during training.

As illustrated in the figure~\ref{fig:realworld-exp}, the three experimental tasks are defined as follows:

The robot moves straight into a circular dark room, exits through the door on the right-hand side, and continues forward until reaching the vicinity of a table.
Starting near the table, the robot proceeds forward while following the curtain on its left-hand side, turns right at the corner, and continues straight until reaching a glass door.
Starting from the glass door, the robot turns right and moves along the corridor, enters the first door on the left, passes a water dispenser, continues straight to the end of the hallway, turns right, and finally stops near a round table in the corridor.

\subsection{Ablation Studies}
To systematically evaluate the efficacy of the proposed three-stage training pipeline, we conduct comprehensive ablation experiments on both seen and unseen environments, with results summarized in Table~\ref{tab:ablation}. This analysis isolates the empirical contributions of each distinct learning phase to the overall navigation performance.

\noindent\textbf{Effectiveness of DAgger Adaptation.} The baseline model, trained exclusively via SFT on the GN-Matrix dataset, establishes foundational instruction-following capabilities but suffers from severe policy degradation due to compounding errors in out-of-distribution states. Incorporating the intermediate DAgger stage effectively mitigates this covariate shift. By exposing the agent to rollout-induced states and injecting corrective oracle supervision, this phase relaxes policy over-concentration and expands search entropy. Consequently, we observe a substantial boost in both SR and OS, demonstrating enhanced state coverage and online execution stability.

\noindent\textbf{Benefits of Preference-Driven RL.} Building upon the robust initialization provided by DAgger, the terminal DAPO stage further aligns the policy with long-horizon execution objectives. As manifested in Table~\ref{tab:ablation}, ablating this reinforcement learning phase (``w/o DAPO'') leads to a noticeable drop in efficiency. The full pipeline consistently minimizes NE and maximizes SPL across both FPV and BEV-conditioned settings. This confirms that while imitation learning stages anchor basic navigation semantics, the preference-driven RL phase is critical for refining sequential decision-making and optimizing global trajectory efficiency.

\begin{table}[t]
    \caption{\textbf{Ablation study of the proposed training pipeline on the validation split.} The ``w/o'' prefix signifies the ablation of a specific stage from the complete pipeline.}
    \label{tab:ablation}
    \centering
    \setlength{\tabcolsep}{4.4pt}
    \begin{NiceTabular}{lcccccccccccc}
        \CodeBefore
            \columncolor[HTML]{EFEFEF}{2-3}
        \Body
        \toprule
        \multirow{2}{*}{\textbf{Method}} & \multicolumn{2}{c}{\textbf{Observation}} & \multicolumn{5}{c}{\textbf{Metrics on Seen}} & \multicolumn{5}{c}{\textbf{Metrics on Unseen}} \\ 
        \cmidrule(lr){2-3} \cmidrule(lr){4-8} \cmidrule(lr){9-13}
         & \textbf{BEV} & \textbf{FPV} & \textbf{TL} & \textbf{NE} $\downarrow$ & \textbf{OS} $\uparrow$ & \textbf{SR} $\uparrow$ & \textbf{SPL} $\uparrow$ & \textbf{TL} & \textbf{NE} $\downarrow$ & \textbf{OS} $\uparrow$ & \textbf{SR} $\uparrow$ & \textbf{SPL} $\uparrow$ \\ 
        \midrule
        Full Pipeline &  & \checkmark & 5.3 & 4.9 & 49.5 & 46.9 & 45.1 & 5.0 & 5.6 & 43.6 & 38.9 & 37.3 \\ 
        w/o DAgger &  & \checkmark & 3.0 & 6.7 & 27.5 & 26.5 & 25.8 & 4.9 & 5.7 & 42.4 & 37.8 & 36.0 \\ 
        w/o DAPO &  & \checkmark & 5.0 & 5.2 & 45.4 & 41.9 & 40.2 & 3.1 & 6.8 & 29.9 & 28.1 & 27.3 \\ 
        w/o DAgger \& DAPO &  & \checkmark & 4.1 & 6.5 & 28.9 & 26.1 & 24.8 & 4.1 & 6.4 & 32.9 & 29.4 & 28.1 \\ 
        \midrule
        Full Pipeline & \checkmark & \checkmark & 5.2 & 4.3 & 59.4 & 58.7 & 58.6 & 4.0 & 5.8 & 40.2 & 38.5 & 38.2 \\ 
        w/o DAgger & \checkmark & \checkmark & 3.0 & 6.9 & 26.9 & 25.6 & 25.0 & 3.2 & 6.4 & 33.5 & 32.4 & 32.3 \\ 
        w/o DAPO & \checkmark & \checkmark & 4.2 & 5.4 & 48.0 & 47.6 & 47.4 & 3.1 & 6.8 & 30.4 & 27.8 & 26.9 \\
        w/o DAgger \& DAPO & \checkmark & \checkmark & 3.8 & 6.8 & 27.2 & 23.6 & 22.4 & 4.0 & 6.6 & 31.4 & 27.6 & 26.4 \\ 
        \bottomrule
    \end{NiceTabular}
\end{table}

\section{Discussion}
This work presents, to the best of our knowledge, the first unified embodied navigation framework spanning data, simulation, model, and evaluation. By constructing the largest high-fidelity 3DGS-based navigation dataset to date and developing an interactive 3DGS simulator, GN-Matrix provides a scalable foundation for long-horizon embodied navigation.

We further introduce GN-Bench, a new benchmark for evaluating embodied navigation with BEV-based metrics and dynamic 3DGS avatars, enabling more realistic assessment of spatial reasoning and human-robot interaction. Built on this framework, BAE combines supervised learning, DAgger, and reinforcement learning to improve robustness under rollout-induced distribution shifts, while 3DGS-rendered BEV serves as an efficient spatial memory for unified map-based and map-free navigation.

Extensive experiments show that GN-BAE achieves state-of-the-art performance on both VLN-CE and GN-Bench. Beyond simulation, it transfers effectively to real robots and supports cross-embodiment adaptation across humanoid and wheeled platforms, highlighting its practical potential for general embodied intelligence.

Despite these advances, incorporating richer physical interactions and extending the framework toward more complex embodied tasks remain important future directions.

\appendix

\newcommand{\pluseq}{\mathrel{+}=}

\section{Additional Details of Data Generation}
\label{app:datagen_details}

This appendix provides additional technical details for the data
generation process of NavGMatrix, including scene-source breakdown,
dynamic human avatar construction, task synthesis details, benchmark
compatibility adaptation, and expanded dataset statistics.

\subsection{Detailed Scene Source Breakdown and Category Taxonomy}
\label{app:scene_breakdown}

In this section, we provide a more detailed breakdown of the scene
assets used to construct NavGMatrix. In particular, we summarize the
contributions of InteriorGS, handcrafted large-scale scenes, and
WorldGrow-expanded scenes to the overall dataset.

We further categorize scenes by environment type and report additional
statistics such as scene counts, layout complexity, room count, and
representative examples. These statistics complement the high-level
scene description in the main paper and help illustrate the scale and
diversity of the underlying 3DGS scene corpus.

\begin{table*}[h]
    \centering
    \caption{Detailed scene-source breakdown for GN-Matrix.}
    \label{tab:scene_source_stats}
    \resizebox{\textwidth}{!}{
    \begin{tabular}{lccccc}
        \toprule
        \textbf{Source} &
        \textbf{\# Scenes} &
        \textbf{Main Categories} &
        \textbf{Avg. Scene Size (m$^2$)} &
        \textbf{Avg. Object Count} &
        \textbf{Notes} \\
        \midrule
        InteriorGS &
        1,000 &
        Indoor residential &
        150.1 &
        502.9 &
        High-quality indoor 3DGS scenes with semantic annotations \\
        Handcrafted large-scale scenes &
        150 &
        Large-scale commercial (office/supermarket) &
        374.5 &
        2321.4 &
        Dense large-scale environments such as offices and supermarkets \\
        WorldGrow-expanded scenes &
        1760 / 240 &
        6x6 and 8x8 indoor scenes &
        91.68 / 158.76 &
        N/A &
        Automatically expanded scenes for improving diversity and scale \\
        \bottomrule
    \end{tabular}}
\end{table*}

\subsubsection{WorldGrow Scene Generation Prompts}
\label{app:worldgrow_prompts}

To expand the scene distribution beyond curated indoor scans, we use
text-conditioned scene generation for WorldGrow-expanded environments.
An example prompt used for generating a high-detail residential scene is
shown below:
\begin{verbatim}
A photorealistic 3D residence mesh with a Victorian Vintage style,
featuring ornate wood paneling and velvet-upholstered seating areas
that lavishly incorporate tufted sofas, carved armchairs, roll-top
desks, four-poster beds, and display cabinets with intricate woodwork.
Use crystal chandeliers and stained glass lamps to create rich, layered
lighting. Prioritize detailed topology for decorative surfaces and
hand-carved wooden components for opulence, maintaining a luxurious and
ornate historical aesthetic.
\end{verbatim}

These prompts are designed to control both global scene style and local
object richness, thereby improving the diversity, density, and visual
fidelity of automatically expanded environments.

\subsection{Dynamic Human Avatar Construction Details}
\label{app:human_avatar_details}

\paragraph{Single-image avatar reconstruction.}
For each input human image, we recover a parametric SMPL-X body model
using SMPLify-X and reconstruct a canonical 3D Gaussian avatar in
T-pose using LHM. This produces two complementary representations: a
structured body prior and a visually faithful Gaussian avatar.

\paragraph{Mesh--Gaussian alignment and binding.}
We align the SMPL-X mesh to the canonical Gaussian avatar and construct
an aligned SMPL-X--Gaussian pair that allows avatar appearance to be
driven by body motion. We consider two fitting strategies:
(1) nearest-neighbor pairing with ICP-style refinement, and
(2) a robust statistical fitting formulation based on Mahalanobis
distance.

\paragraph{Animation retargeting and motion driving.}
We collect motion assets in FBX format from Mixamo and convert them
into frame-wise SMPL-X parameters. These per-frame parameters are then
used to drive the aligned SMPL-X--Gaussian pair and produce animated
per-frame Gaussian avatar sequences.

\paragraph{Avatar and motion statistics.}
We additionally summarize the scale and diversity of the dynamic human
avatar component, including the number of reconstructed identities,
appearance keyword coverage, motion categories, and animation counts.

\subsubsection{Problem setup}
Let the SMPL-X body model be parameterized by shape $\beta$ and pose
$\theta$, with vertices and joints denoted by
$V(\beta,\theta)=\{v_i\}_{i=1}^{N_V}$ and
$J(\beta,\theta)=\{j_m\}_{i=1}^{N_J}$, respectively. After applying a
global alignment transform, the transformed source points are denoted
by
\[
\tilde{v}_i = R_{\text{pre}} (R v_i + t), \qquad
\tilde{j}_m = R_{\text{pre}} (R j_m + t),
\]
where $R \in SO(3)$ is the current global rotation, $t \in \mathbb{R}^3$
is the global translation, and $R_{\text{pre}} \in SO(3)$ is an
optional fixed pre-rotation.

Let the canonical Gaussian avatar be represented by a set of Gaussian
components
\[
\mathcal{G} = \{(\mu_k,\Sigma_k,\alpha_k)\}_{k=1}^{N_G},
\]
where $\mu_k \in \mathbb{R}^3$ is the Gaussian center,
$\Sigma_k \in \mathbb{R}^{3\times 3}$ is the covariance matrix,
and $\alpha_k \in [0,1]$ is the opacity or confidence weight. We write
the corresponding precision matrix as
\[
\Lambda_k = \Sigma_k^{-1}.
\]

\subsubsection{ICP-style nearest-neighbor fitting}
Given a transformed source point $x$, we first retrieve a candidate
neighbor set $\mathcal{N}(x)$ from the Gaussian centers using Euclidean
distance. A soft correspondence centroid is then defined as
\[
q(x)=\sum_{k\in\mathcal{N}(x)} w_k(x)\mu_k,
\qquad
w_k(x)=
\frac{\alpha_k \exp\!\left(-\frac{\|x-\mu_k\|_2^2}{\sigma_d^2}\right)}
{\sum_{\ell\in\mathcal{N}(x)}
\alpha_\ell \exp\!\left(-\frac{\|x-\mu_\ell\|_2^2}{\sigma_d^2}\right)}.
\]

Using the resulting source--target pairs $(x,q(x))$, we form a trimmed
rigid alignment objective
\[
\mathcal{L}_{\text{rigid}}=
\frac{1}{|\mathcal{S}_{\text{trim}}|}
\sum_{x\in\mathcal{S}_{\text{trim}}}
\omega(x)\,\rho\!\left(\|x-q(x)\|_2\right),
\]
where $\mathcal{S}_{\text{trim}}$ denotes the retained source set after
trimming large-residual correspondences, $\omega(x)$ is a scalar
correspondence confidence, and $\rho(\cdot)$ is a robust loss such as
the Huber loss.

\subsubsection{Robust Mahalanobis fitting}
To better account for anisotropic Gaussian geometry, we also consider a
hard Mahalanobis assignment objective. For each source point $x$, we
choose the best-matching Gaussian index
\[
k^\star(x)=\arg\min_{k\in\mathcal{N}(x)}
(x-\mu_k)^\top \Lambda_k (x-\mu_k).
\]
The corresponding robust fitting loss is
\[
\mathcal{L}_{\text{maha}}=
\frac{1}{|\mathcal{S}_{\text{trim}}|}
\sum_{x\in\mathcal{S}_{\text{trim}}}
\rho\!\left(
\sqrt{(x-\mu_{k^\star(x)})^\top
\Lambda_{k^\star(x)}
(x-\mu_{k^\star(x)})}
\right).
\]

\subsubsection{Optional soft-assignment refinement}
As an alternative soft correspondence formulation, we may define a
Gaussian mixture assignment around each source point. The soft weight of
Gaussian $k$ for source point $x$ is
\[
\pi_k(x)=
\frac{\alpha_k
\exp\!\left(
-\frac{1}{T}
(x-\mu_k)^\top \Lambda_k (x-\mu_k)
\right)}
{\sum_{\ell\in\mathcal{N}(x)} \alpha_\ell
\exp\!\left(
-\frac{1}{T}
(x-\mu_\ell)^\top \Lambda_\ell (x-\mu_\ell)
\right)},
\]
where $T$ is a temperature parameter. The corresponding soft
Mahalanobis loss can be written as
\[
\mathcal{L}_{\text{soft}}=
\frac{1}{|\mathcal{S}|}
\sum_{x\in\mathcal{S}}
\sum_{k\in\mathcal{N}(x)}
\pi_k(x)\,
\rho\!\left(
\sqrt{(x-\mu_k)^\top \Lambda_k (x-\mu_k)}
\right).
\]

\subsubsection{Anchor and regularization terms}
To stabilize alignment, we may additionally use anchor constraints on
selected joints:
\[
\mathcal{L}_{\text{anchor}}=
\frac{1}{|\mathcal{A}|}
\sum_{m\in\mathcal{A}}
\operatorname{Huber}\!\left(
\|\tilde{j}_m-a_m\|_2;\delta
\right),
\]
where $\mathcal{A}$ is an anchor joint set and $a_m$ is the target
anchor position for joint $m$.

A regularized pose refinement objective can then be written as
\[
\mathcal{L}_{\text{fit}}=
\lambda_{\text{rigid}}\mathcal{L}_{\text{rigid}}
+
\lambda_{\text{maha}}\mathcal{L}_{\text{maha}}
+
\lambda_{\text{soft}}\mathcal{L}_{\text{soft}}
+
\lambda_{\text{anchor}}\mathcal{L}_{\text{anchor}}
+
\lambda_{\text{prior}}\mathcal{L}_{\text{prior}},
\]
where $\mathcal{L}_{\text{prior}}$ denotes body pose and shape
regularization.

\subsubsection{Avatar animation pipeline}
Given the aligned SMPL-X--Gaussian pair, we convert motion sequences
from Mixamo FBX assets into frame-wise SMPL-X parameters
$\{\theta_t\}_{t=1}^{T}$. For each frame $t$, the corresponding body
parameters are applied to the aligned SMPL-X model, and the resulting
motion is transferred to the Gaussian avatar through the learned
binding relationship. This yields an animated per-frame Gaussian avatar
sequence
\[
\{\mathcal{G}_t\}_{t=1}^{T}.
\]

\begin{table}[h]
    \centering
    \caption{Suggested appendix table for avatar and motion statistics.}
    \label{tab:avatar_motion_stats}
    \begin{tabular}{p{0.28\linewidth} p{0.22\linewidth} p{0.32\linewidth}}
        \toprule
        \textbf{Category} & \textbf{Used Count} & \textbf{Notes} \\
        \midrule
        Human identities & & Reconstructed avatar count \\
          - SHHQ identities & 76 & Source breakdown \\
          - HuGe100K identities & 100 & Source breakdown \\
        Motion categories & 4 & Walking / Running / Waving / Idle \\
        \bottomrule
    \end{tabular}
\end{table}

\subsection{Detailed Task Generation Algorithms}
\label{app:task_algorithms}

\subsubsection{Goal-Based Navigation Task Generation}
\label{app:goal_task_details}

\paragraph{Task formulation and target construction.}
For each scene, we first construct a candidate target pool from the
available semantic scene annotations. Each candidate target corresponds
to a meaningful object instance or target region that can serve as a
navigation goal. Since raw annotations may contain nearby duplicated
instances, heavily overlapping targets, or semantically redundant
endpoints, we further apply redundancy suppression to obtain a cleaner
goal set for task generation.

The purpose of this step is to ensure that the resulting goal-based
navigation tasks are grounded in semantically meaningful scene targets
rather than arbitrary endpoints. This improves the interpretability of
the generated tasks and strengthens compatibility with downstream
instruction generation and benchmark conversion.

\paragraph{Implementation-aligned target construction.}
\begin{algorithm}[h]
\caption{Goal Target Construction and Safe Goal Snapping}
\label{alg:goal_target_construction}
\begin{algorithmic}[1]
\Require Occupancy map, occupancy metadata, structure annotations, semantic labels
\Ensure Filtered semantic target set $\mathcal{T}$
\State Load grayscale occupancy image and optional wall mask
\State Compute traversable grid by thresholding occupancy values
\State Inflate obstacles with safety margin to obtain safe grid $\mathcal{G}_{\text{safe}}$
\State Load wall segments from scene structure annotations
\State Compute connected components on $\mathcal{G}_{\text{safe}}$
\State $\mathcal{T}\gets\emptyset$
\For{each semantic instance $e$ in \texttt{labels.json}}
    \State Compute object centroid $(c_x,c_y,c_z)$ from its bounding box
    \State Convert centroid to pixel location $g_{\text{raw}}$
    \State $g\gets\textsc{FindNearestSafe}(g_{\text{raw}}, \mathcal{G}_{\text{safe}}, \text{walls})$
    \If{$g$ is invalid} \textbf{continue} \EndIf
    \If{$g$ does not belong to any valid connected component} \textbf{continue} \EndIf
    \State Add target record
    \[
    \{ \texttt{ins\_id}, \texttt{label}, \texttt{centroid\_world}, \texttt{goal\_pixel}, \texttt{goal\_component} \}
    \]
    \State to $\mathcal{T}$
\EndFor
\State \Return $\mathcal{T}$
\end{algorithmic}
\end{algorithm}

We do not merge object instances via IoU at the target construction
stage. Instead, near-duplicate endpoints are suppressed later through
path-level similarity filtering, which better preserves semantic target
coverage while removing redundant episodes.

\paragraph{Start--goal sampling and reference path generation.}
After constructing the filtered target pool, we sample feasible
start--goal pairs over the traversable region of the scene. Start
positions are sampled from valid navigable space, while goal positions
are grounded in semantic targets. For each feasible pair, we generate a
reference trajectory over the scene traversability representation to
obtain a navigation path from the sampled start state to the target.

To avoid excessive concentration around a small number of spatial
regions, the sampling process is designed to encourage broader spatial
coverage across the scene. This improves task diversity and produces a
wider range of path lengths and spatial configurations.

\paragraph{Planning and diversification algorithm.}
For each cleaned semantic target, we generate a reference trajectory on the inflated safe occupancy grid. The target location is fixed to its snapped safe-grid goal, and the start is sampled
from the same connected component to guarantee reachability. To improve spatial coverage, the sampler uses inverse-density zone weighting when scene-level visitation statistics are available,
otherwise biases sampling toward locations far from previously used points, and falls back to uniform component sampling when no coverage prior is provided.

The sampled start and target goal are connected with A* search on the safe grid using local grid connectivity and obstacle-aware traversal. Invalid searches are discarded. Valid trajectories
are then simplified by line-of-sight smoothing, densified with rasterized line segments, and adaptively resampled into compact keypoints. Finally, all geometric annotations are converted from
pixel space to world space, and the initial facing direction is estimated from the first trajectory segment, yielding a target-grounded navigation record with start, goal, keypoints, dense
trajectory, and metadata.
\begin{algorithm}[t]
\caption{Goal-Conditioned Reference Trajectory Generation}
\label{alg:goal_conditioned_reference_generation}
\begin{algorithmic}[1]
\Require Safe grid $\mathcal{G}_{\text{safe}}$, connected components $\mathcal{C}$, target set $\mathcal{T}$, optional coverage bias
\Ensure Generated path set $\mathcal{P}$
\State $\mathcal{P}\gets\emptyset$
\For{each target $\tau\in\mathcal{T}$}
    \State $g\gets\tau.\texttt{goal\_pixel}$
    \State $c\gets\tau.\texttt{goal\_component}$
    \If{zone statistics exist}
        \State Sample start pixel $s$ from component $c$ using inverse-density zone weighting
    \ElsIf{coverage bias points exist}
        \State Sample start pixel $s$ from component $c$ to maximize distance from prior sampled points
    \Else
        \State Uniformly sample start pixel $s$ from component $c$
    \EndIf
    \State $p\gets\textsc{AStar}(\mathcal{G}_{\text{safe}}, s, g)$
    \If{$p$ is invalid} \textbf{continue} \EndIf
    \State Smooth $p$ using line-of-sight simplification
    \State Densify the smoothed path using rasterized line segments
    \State Adaptively resample keypoints along the path
    \State Convert start, goal, raster path, and keypoints from pixel space to world space
    \State Estimate start-facing direction from the first path segment
    \State Emit one path record with semantic target, start, goal, keypoints, raster path, and metadata
    \State $\mathcal{P}\gets\mathcal{P}\cup\{p\}$
\EndFor
\State \Return $\mathcal{P}$
\end{algorithmic}
\end{algorithm}

\paragraph{Trajectory validity and diversity filtering.}
Not all generated reference trajectories are retained in the final
dataset. To improve overall data quality, we apply a series of validity
and diversity filters to remove undesirable episodes. In particular, we
reject trajectories that are too short, spatially trivial, highly
similar to previously accepted paths, or insufficiently distinct in
their start--goal configuration.

These filters reduce the number of near-duplicate or low-information
episodes and help the final dataset better represent long-horizon,
diverse navigation behavior. As a result, the accepted task set is more
balanced in both path geometry and spatial coverage.

\paragraph{Post-generation filtering algorithm.}
After generation, candidate trajectories are filtered before being retained in the final task set. We first remove malformed records and trajectories whose geometric length falls below the
minimum threshold. We then reject exact duplicates using stored trajectory signatures, ensuring that previously accepted data are not regenerated as identical samples.

To suppress near-duplicates, each remaining candidate is compared against both prior accepted trajectories and trajectories already kept in the current generation pass. Candidates are only
considered comparable when their start and goal locations are within fixed spatial thresholds; for such cases, the discrete Fr\'echet distance is computed on resampled trajectories. A
candidate is rejected if this distance falls below the similarity threshold. This post-generation filtering step improves task diversity by reducing repeated start--goal configurations,
short-horizon episodes, and geometrically redundant trajectories while preserving valid long-range navigation behavior.
\begin{algorithm}[t]
\caption{Post-Generation Validity and Diversity Filtering}
\label{alg:post_generation_filtering}
\begin{algorithmic}[1]
\Require Generated path set $\mathcal{P}$, existing accepted paths $\mathcal{P}_{\text{old}}$
\Ensure Filtered path set $\mathcal{P}_{\text{keep}}$
\State Load all prior accepted paths and their signatures
\State $\mathcal{P}_{\text{keep}}\gets\emptyset$
\For{each candidate path $p\in\mathcal{P}$}
    \State Parse path record and compute path length
    \If{$p$ is unreadable} \textbf{continue} \EndIf
    \If{path length is below minimum threshold} \textbf{continue} \EndIf
    \If{signature of $p$ already exists} \textbf{continue} \EndIf
    \State $\texttt{similar}\gets\textbf{False}$
    \For{each prior path $q\in\mathcal{P}_{\text{old}}\cup\mathcal{P}_{\text{keep}}$}
        \If{start distance between $p$ and $q$ exceeds threshold} \textbf{continue} \EndIf
        \If{goal distance between $p$ and $q$ exceeds threshold} \textbf{continue} \EndIf
        \State Compute discrete Fr\'echet distance between resampled versions of $p$ and $q$
        \If{distance is below similarity threshold}
            \State $\texttt{similar}\gets\textbf{True}$
            \State \textbf{break}
        \EndIf
    \EndFor
    \If{$\texttt{similar}$} \textbf{continue} \EndIf
    \State Accept $p$
    \State $\mathcal{P}_{\text{keep}}\gets\mathcal{P}_{\text{keep}}\cup\{p\}$
\EndFor
\State \Return $\mathcal{P}_{\text{keep}}$
\end{algorithmic}
\end{algorithm}

\paragraph{Generated supervision and export format.}
Each accepted trajectory is packaged into a goal-based navigation task
with semantic target information, reference path annotations, and
downstream supervision targets. Depending on the export setting, the
task may further provide action labels, waypoint or trajectory
supervision, first-person observations, bird's-eye-view context, and
benchmark-compatible instruction formats.

This packaging design allows the same underlying goal-based navigation
task to support multiple downstream training and evaluation settings,
including both native NavGMatrix supervision and converted benchmark
formats.

\subsubsection{Human-Following Task Generation}
\label{app:human_follow_details}

We construct human-following episodes by reusing feasible goal-navigation
paths and converting them into time-parameterized human motion. For each
feasible path, we sample a motion category (e.g., cruise, turn,
stop-and-go), retime the path with bounded speed and turn rate, and
instantiate an animated human avatar along this trajectory. The agent is
initialized behind the human with a preferred offset and is required to
track a moving reference pose while preserving distance and viewpoint
constraints.

Let the human state be $x_h^t=(p_h^t,\psi_h^t)$ and agent state be
$x_a^t=(p_a^t,\psi_a^t)$ at time $t$. The preferred follow pose is
\[
p_{\text{ref}}^t = p_h^t - d_{\text{pref}}\hat f_h^t + \ell \hat r_h^t,\quad
\psi_{\text{ref}}^t = \mathrm{atan2}(p_h^t - p_a^t),
\]
where $d_{\text{pref}}\in[d_{\min},d_{\max}]$ is the follow distance and
$\ell\in[-\ell_{\max},\ell_{\max}]$ is the lateral offset.

\begin{algorithm}[t]
\caption{Compact Human-Following Episode Generation}
\label{alg:human_follow_compact}
\begin{algorithmic}[1]
\Require Feasible nav paths $\mathcal{P}_{\text{nav}}$, map $\mathcal{M}$, params $\Theta$
\Ensure Valid follow set $\mathcal{E}$
\State $\mathcal{E}\gets\emptyset$
\For{$P\in\mathcal{P}_{\text{nav}}$}
  \State $c\sim p(c)$; $X_h\gets\textsc{RetimePath}(P,c,\Theta)$
  \If{$\textsc{InvalidHumanPath}(X_h,\mathcal{M})$} \textbf{continue} \EndIf
  \State $(d_{\text{pref}},\ell)\gets\textsc{SampleOffsets}(\Theta)$
  \State $x_a^0\gets\textsc{SpawnBehind}(x_h^0,d_{\text{pref}},\ell,\mathcal{M})$
  \If{spawn invalid} \textbf{continue} \EndIf
  \State violations $\gets 0$
  \For{$t=1,\dots,T$}
    \State $x_{\text{ref}}^t\gets\textsc{FollowRef}(x_h^t,d_{\text{pref}},\ell)$
    \State $x_a^t\gets\textsc{TrackerStep}(x_a^{t-1},x_{\text{ref}}^t,\mathcal{M},\Theta)$
    \State violations $\pluseq\textsc{Check}(x_a^t,x_h^t,\Theta)$
  \EndFor
  \If{$\textsc{Reject}(violations,\Theta)$} \textbf{continue} \EndIf
  \State $\mathcal{E}\gets\mathcal{E}\cup\{(X_h,X_a,c)\}$
\EndFor
\State \Return $\mathcal{E}$
\end{algorithmic}
\end{algorithm}

\subsection{Instruction Generation and Benchmark Adaptation Details}
\label{app:instruction_generation}

To support compatibility with multiple embodied navigation benchmarks,
we adapt NavGMatrix tasks into benchmark-specific instruction and
action-space formats. This conversion preserves the original spatial and
semantic structure of the generated tasks while making them usable in
existing training and evaluation pipelines.

\paragraph{Instruction adaptation.}
Task specifications are reformulated into benchmark-compatible
instruction styles aligned with VLN-CE, VLN-PE, and NaVid. Depending on
the benchmark, this may involve converting semantic targets and
reference trajectories into instruction-oriented task descriptions while
preserving the original navigation objective.

\paragraph{Discrete action-space adaptation.}
In addition to instruction adaptation, the generated reference
trajectories are converted into benchmark-compatible discrete action
sequences. This allows NavGMatrix to support policy-learning settings
that operate on discrete control spaces rather than continuous
trajectory representations.

\paragraph{Path generation: moving instruction.}
For path generation, we use a prompt that converts structured segment
summaries into concise movement instructions:
\begin{verbatim}
System prompt:
You are a robotics navigation instructor. Based on the path summary,
craft a concise English instruction. Describe straight segments using
'go straight for <distance> meters' without referencing heading angles.
When the summary indicates a significant turn, include an explicit
'turn left' or 'turn right'. Avoid words like 'slightly'.

User prompt:
Scene: <scene_id>
Target object: <label_name>
Path summary (follow in order):
<segment_summary_lines>
Return JSON exactly as {"moving_instruction": "..."}.
\end{verbatim}

\paragraph{Base instruction minimal grammar fix.}
For concise base instructions, we apply a lightweight editing prompt
that performs minimal grammar correction without changing the navigation
semantics:
\begin{verbatim}
System prompt:
You are an editor. Given a navigation instruction, make the smallest
possible grammar fix to correct articles/fluency (e.g., 'an other' ->
'another', 'get a water' -> 'get some water'). Do not add extra context
or change the meaning. If it is already fine, return it unchanged.
Return JSON exactly as {"instruction": "..."}.

User prompt:
Scene: <scene_id>
Target object/place: <label_name>
Instruction: <instruction_text>
\end{verbatim}

\paragraph{Detailed and benchmark-style instruction generation.}
For more detailed route descriptions, we generate fluent instructions
from structured step breakdowns. This prompt also serves as the basis
for benchmark-style instruction generation when route descriptions must
be aligned with object- or landmark-aware annotation conventions:
\begin{verbatim}
System:
You are good at guiding the way. Generate fluent, step-by-step
navigation instructions from structured inputs.

Rules:
- Keep the action order exactly as provided; never add or reorder steps.
- Select one best scene tag or landmark per step. For move/advance
  actions, prefer areas/rooms; for turns, prefer specific objects near
  the turn.
- Use natural, concise English with smooth transitions (then, next,
  after that) and no explicit numbering or the word "step".
- Do not mention distances or coordinates.
- Mention the target only once, at the end, as the arrival point.

Style hints:
- Forward-facing point of view; describe turns as left/right relative
  to facing direction.
- Reference distinctive objects rather than structural elements
  (walls, ceilings) unless nothing else is available.
- Keep to one or two sentences total, covering all moves and turns.

Template:
{style_text}

Scene: {scene_id}
Goal object: {goal_label}
Step breakdown:
{step_breakdown}

Write the grounded instruction now.
\end{verbatim}

\paragraph{Grounded instruction review and refinement.}
To improve faithfulness and linguistic quality, we further refine the
generated landmark-aware instruction with a review prompt:
\begin{verbatim}
You are editing a navigation instruction. Keep the step order and count
identical; do not add steps or landmarks that are not provided.
Use at most <max_landmarks> distinct landmarks in total; avoid repeating
the same landmark unless necessary.
Prefer larger, distinctive landmarks; avoid ceiling/floor or tiny
clutter. If landmarks are weak, keep the instruction brief.
Output a single concise paragraph (no numbering or bullet points). Do
not mention unavailable objects.

Goal object: <goal_label>
Step breakdown:
<step_breakdown_lines>

Available landmarks (deduped): <comma_separated_landmarks_or_none>

Raw instruction to refine:
<raw_instruction>

Rewrite now. Respond with the instruction only.
\end{verbatim}

\paragraph{Instruction style taxonomy.}
We use three instruction styles in NavGMatrix:
\begin{itemize}
    \item \textbf{Base instruction.} A short and simple target-oriented
    instruction used as the most concise natural-language form.
    \item \textbf{Detailed instruction.} A more descriptive instruction
    generated from structured step breakdowns and action sequences.
    \item \textbf{Benchmark-style instruction.} A landmark-aware
    instruction aligned with conventions such as VLN-CE and VLN-PE,
    where the route is described with reference items or scene
    landmarks.
\end{itemize}

\subsection{Expanded Dataset Statistics and Balancing Details}
\label{app:expanded_statistics}

In this section, we report expanded statistics for NavGMatrix,
including task composition, path distributions, and action balancing.

\paragraph{Path and task statistics.}
We summarize additional statistics such as path length distributions,
per-scene episode counts, static versus dynamic task proportions, and
object-category frequency.

\paragraph{Action balancing.}
Since the raw generated dataset is strongly skewed toward forward
actions, we rebalance the action distribution to improve downstream
policy learning. We report the action distribution before and after
balancing in the appendix.

\begin{table}[t]
\centering
\small
\caption{Action distribution used for dataset balancing. The balanced dataset increases the relative proportion of turning actions while preserving a smaller but meaningful proportion of stop actions.}
\label{tab:action_balance_appendix}
\begin{tabular}{l>{\columncolor{gray!15}}c c}
\toprule
\textbf{Action} & \textbf{Percentage W/ Adjustment} & \textbf{Percentage W/O Adjustment} \\
\midrule
Forward     & 67\%  & 85.6\% \\
Turn left   & 15\%  & 6.9\% \\
Turn right  & 15\%  & 6.7\% \\
Stop        & 3\%   & 0.8\% \\
\midrule
Total       & 100\% & 100\% \\
\bottomrule
\end{tabular}
\end{table}

\subsection{Additional Qualitative Examples}
\label{app:qualitative_examples}

We include additional qualitative examples to illustrate the diversity
and difficulty of the generated data beyond the examples shown in the
main paper.

\begin{itemize}
    \item Additional examples of goal-based navigation trajectories.
    \item Additional examples of human-following episodes.
    \item Examples under different visual conditions.
    \item Challenging scenes, failure cases, and edge cases.
\end{itemize}

\subsection{Model Details}
\subsubsection{VLM prompts}

We provide the raw prompt texts used in our VLM pipeline below.
The planning task uses three prompt variants, while the following task
uses two variants. In implementation, the following task is instantiated
as \texttt{TRACKING}.

\paragraph{Planning prompt with RGB, history, BEV, and OCC.}

\begin{quote}
\small
\texttt{<image><image><image><image>}\\
You are given FOUR images from the same navigation episode.\\
\\
Image 1: CURRENT first-person RGB.\\
Image 2: HISTORY mosaic (4x4), NEW->OLD sampled by discrete action steps (current excluded; missing => black).\\
Image 3: Scene-level BEV map.\\
Image 4: OCC / traversability map (ORIGINAL resolution). Green=start, Red=current, white=traversable.\\
\\
TASK\_TYPE: PLANNING\\
Instruction: "\{INSTRUCTION\_TEXT\}"\\
\\
COORDS (Image 4, normalized tokens):\\
- Each coordinate is (<x\_norm>,<y\_norm>) as tokens <0>..<999>.\\
- OCC size: W=\{OCC\_W\} px, H=\{OCC\_H\} px\\
- Normalize: x\_norm = round\_half\_up(x\_px/W*1000), y\_norm = round\_half\_up(y\_px/H*1000), clip [0,999].\\
- Scale: 1 x\_norm = \{X\_NORM\_UNIT\_M\} m, 1 y\_norm = \{Y\_NORM\_UNIT\_M\} m. (1 px = \{OCC\_M\_PER\_PX\} m)\\
\\
CURRENT\_PIXEL (normalized): (\{CUR\_XN\},\{CUR\_YN\})\\
\\
Your task (return all branches aligned):\\
(A) Waypoints on Image 4, normalized tokens, 13 points: current + 12 future.\\
\hspace*{1em}- Point0 = CURRENT\_PIXEL.\\
\hspace*{1em}- Points1..12 follow future path with 0.25m spacing (3.0m total); pad with last if short.\\
(B) Actions: exactly 6 tokens from \{\{<FWD>,<LEFT>,<RIGHT>,<STOP>\}\} matching (A).\\
\hspace*{1em}- Action semantics: <FWD> moves forward 0.25m; <LEFT>/ <RIGHT> rotate 15 degrees in place (no translation); <STOP> means stop.\\
\\
Output (STRICT): JSON with keys "pixel" and "vlnce".\\
- "pixel": "[[<x0>,<y0>],...,(13 pts)]"\\
- "vlnce": "<action><FWD>,...,<STOP></action>"\\
No extra text.
\end{quote}

\paragraph{Planning prompt with BEV and OCC.}
\begin{quote}
\small
\texttt{<image><image>}\\
You are given TWO images from the same navigation episode.\\
\\
Image 1: Scene-level BEV map.\\
Image 2: OCC / traversability map (ORIGINAL resolution). Green=start, Red=current, white=traversable.\\
\\
TASK\_TYPE: PLANNING\\
Instruction: "\{INSTRUCTION\_TEXT\}"\\
\\
COORDS (Image 2, normalized tokens):\\
- Each coordinate is (<x\_norm>,<y\_norm>) as tokens <0>..<999>.\\
- OCC size: W=\{OCC\_W\} px, H=\{OCC\_H\} px\\
- Normalize: x\_norm = round\_half\_up(x\_px/W*1000), y\_norm = round\_half\_up(y\_px/H*1000), clip [0,999].\\
- Scale: 1 x\_norm = \{X\_NORM\_UNIT\_M\} m, 1 y\_norm = \{Y\_NORM\_UNIT\_M\} m. (1 px = \{OCC\_M\_PER\_PX\} m)\\
\\
CURRENT\_PIXEL (normalized): (\{CUR\_XN\},\{CUR\_YN\})\\
\\
Your task (return all branches aligned):\\
(A) Waypoints on Image 2, normalized tokens, 13 points: current + 12 future.\\
\hspace*{1em}- Point0 = CURRENT\_PIXEL.\\
\hspace*{1em}- Points1..12 follow future path with 0.25m spacing (3.0m total); pad with last if short.\\
(B) Actions: exactly 6 tokens from \{\{<FWD>,<LEFT>,<RIGHT>,<STOP>\}\} matching (A).\\
\hspace*{1em}- Action semantics: <FWD> moves forward 0.25m; <LEFT>/ <RIGHT> rotate 15 degrees in place (no translation); <STOP> means stop.\\
\\
Output (STRICT): JSON with keys "pixel" and "vlnce".\\
- "pixel": "[[<x0>,<y0>],...,(13 pts)]"\\
- "vlnce": "<action><FWD>,...,<STOP></action>"\\
No extra text.
\end{quote}

\paragraph{Planning prompt with RGB and history only.}
\begin{quote}
\small
\texttt{<image><image>}\\
You are given TWO images from the same navigation episode.\\
\\
Image 1: CURRENT first-person RGB.\\
Image 2: HISTORY mosaic (4x4), NEW->OLD sampled by discrete action steps (current excluded; missing => black).\\
\\
TASK\_TYPE: PLANNING\\
Instruction: "\{INSTRUCTION\_TEXT\}"\\
\\
Your task:\\
- Actions: exactly 6 tokens from \{\{<FWD>,<LEFT>,<RIGHT>,<STOP>\}\}.\\
- Action semantics: <FWD> moves forward 0.25m; <LEFT>/ <RIGHT> rotate 15 degrees in place (no translation); <STOP> means stop.\\
\\
Output (STRICT): JSON with key "vlnce".\\
- "vlnce": "<action><FWD>,...,<STOP></action>"\\
No extra text.
\end{quote}

\paragraph{Following prompt with FPV, BEV, and OCC.}
\begin{quote}
\small
\texttt{<image><image><image><image>}\\
You are given FOUR images from the same navigation episode.\\
\\
Image 1: CURRENT first-person RGB.\\
Image 2: HISTORY mosaic (4x4), ALL BLACK for tracking.\\
Image 3: Scene-level BEV map.\\
Image 4: OCC / traversability map (ORIGINAL resolution). Green=start, Red=current, white=traversable.\\
\\
TASK\_TYPE: TRACKING\\
Instruction: "\{INSTRUCTION\_TEXT\}"\\
\\
COORDS (Image 4, normalized tokens):\\
- Each coordinate is (<x\_norm>,<y\_norm>) as tokens <0>..<999>.\\
- OCC size: W=\{OCC\_W\} px, H=\{OCC\_H\} px\\
- Normalize: x\_norm = round\_half\_up(x\_px/W*1000), y\_norm = round\_half\_up(y\_px/H*1000), clip [0,999].\\
- Scale: 1 x\_norm = \{X\_NORM\_UNIT\_M\} m, 1 y\_norm = \{Y\_NORM\_UNIT\_M\} m. (1 px = \{OCC\_M\_PER\_PX\} m)\\
\\
CURRENT\_PIXEL (normalized): (\{CUR\_XN\},\{CUR\_YN\})\\
\\
Your task (return both branches aligned):\\
(A) Waypoints on Image 4, normalized tokens, 7 points: current + 6 future.\\
\hspace*{1em}- Point0 = CURRENT\_PIXEL; pad with last if short.\\
(B) Actions: exactly 6 tokens from \{\{<FWD>,<LEFT>,<RIGHT>,<STOP>\}\} matching (A).\\
\hspace*{1em}- Action semantics: <FWD> moves forward 0.25m; <LEFT>/ <RIGHT> rotate 15 degrees in place (no translation); <STOP> means stop.\\
\\
Output (STRICT): JSON with keys "pixel" and "vlnce".\\
- "pixel": "[[<x0>,<y0>],...,(7 pts)]"\\
- "vlnce": "<action><FWD>,...,<STOP></action>"\\
No extra text.
\end{quote}

\paragraph{Following prompt with FPV only.}
\begin{quote}
\small
\texttt{<image><image>}\\
You are given TWO images from the same navigation episode.\\
\\
Image 1: CURRENT first-person RGB.\\
Image 2: HISTORY mosaic (4x4), ALL BLACK for tracking.\\
\\
TASK\_TYPE: TRACKING\\
Instruction: "\{INSTRUCTION\_TEXT\}"\\
\\
Your task:\\
- Actions: exactly 6 tokens from \{\{<FWD>,<LEFT>,<RIGHT>,<STOP>\}\}.\\
- Action semantics: <FWD> moves forward 0.25m; <LEFT>/ <RIGHT> rotate 15 degrees in place (no translation); <STOP> means stop.\\
\\
Output (STRICT): JSON with key "vlnce".\\
- "vlnce": "<action><FWD>,...,<STOP></action>"\\
No extra text.
\end{quote}

\bibliographystyle{unsrtnat}
\bibliography{references}

@misc{chu2026abotn0technicalreportvla,
  title         = {ABot-N0: Technical Report on the VLA Foundation Model for Versatile Embodied Navigation},
  author        = {Chu, Zedong and Xie, Shichao and Wu, Xiaolong and Shen, Yanfen and Luo, Minghua and Wang, Zhengbo and Liu, Fei and Leng, Xiaoxu and Hu, Junjun and Yin, Mingyang and Lu, Jia and Guo, Yingnan and Yang, Kai and Han, Jiawei and Chen, Xu and Zhu, Yanqing and Zhao, Yuxiang and Liu, Xin and Yang, Yirong and He, Ye and Wang, Jiahang and Cai, Yang and Zhang, Tianlin and Gao, Li and Liu, Liu and Sun, Mingchao and Jiang, Fan and Wang, Chiyu and Liu, Zhicheng and Pan, Hongyu and Han, Honglin and Gu, Zhining and Yang, Kuan and Zhang, Jianfang and Jing, Di and Guan, Zihao and Guo, Wei and Liu, Guoqing and Yang, Di and Yang, Xiangpo and Yang, Menglin and Xing, Hongguang and Li, Weiguo and Xu, Mu},
  year          = {2026},
  eprint        = {2602.11598},
  archivePrefix = {arXiv},
  primaryClass  = {cs.RO}
}

@inproceedings{miao2026towards,
  title     = {Towards Physically Executable 3D Gaussian for Embodied Navigation},
  author    = {Miao, Bingchen and Wei, Rong and Ge, Zhiqi and Sun, Xiaoquan and Gao, Shiqi and Zhu, Jingzhe and Wang, Renhan and Tang, Siliang and Xiao, Jun and Tang, Rui and Li, Juncheng},
  booktitle = {The Fourteenth International Conference on Learning Representations},
  year      = {2026}
}

@inproceedings{wang2026conavbench,
  title     = {CoNavBench: Collaborative Long-Horizon Vision-Language Navigation Benchmark},
  author    = {Wang, Tianhang and Li, Xinhai and Lu, Fan and Gong, Tianshi and Dong, Jiankun and Xue, Weiyi and Qu, Sanqing and Bai, Chenjia and Chen, Guang},
  booktitle = {The Fourteenth International Conference on Learning Representations},
  year      = {2026}
}

@article{cai2025navdp,
  title   = {NavDP: Learning Sim-to-Real Navigation Diffusion Policy with Privileged Information Guidance},
  author  = {Cai, Wenzhe and Peng, Jiaqi and Yang, Yuqiang and Zhang, Yujian and Wei, Meng and Wang, Hanqing and Chen, Yilun and Wang, Tai and Pang, Jiangmiao},
  journal = {arXiv preprint arXiv:2505.08712},
  year    = {2025}
}

@misc{internvla-n1,
    title = {{InternVLA-N1: An} Open Dual-System Navigation Foundation Model with Learned Latent Plans},
    author = {InternNav Team},
    year = {2025},
    booktitle = {arXiv},
}

@inproceedings{savva2019habitat,
  title     = {Habitat: A Platform for Embodied AI Research},
  author    = {Savva, Manolis and Kadian, Abhishek and Maksymets, Oleksandr and Zhao, Yili and Wijmans, Erik and Jain, Bhavana and Straub, Julian and Liu, Jia and Koltun, Vladlen and Malik, Jitendra and others},
  booktitle = {Proceedings of the IEEE/CVF International Conference on Computer Vision},
  pages     = {9339--9347},
  year      = {2019}
}

@article{zhang2024navid,
  title   = {NaVid: Video-based VLM Plans the Next Step for Vision-and-Language Navigation},
  author  = {Zhang, Jiazhao and Wang, Kunyu and Xu, Rongtao and Zhou, Gengze and Hong, Yicong and Fang, Xiaomeng and Wu, Qi and Zhang, Zhizheng and Wang, He},
  journal = {Robotics: Science and Systems},
  year    = {2024}
}

@article{zhang2024uni,
  title   = {Uni-NaVid: A Video-based Vision-Language-Action Model for Unifying Embodied Navigation Tasks},
  author  = {Zhang, Jiazhao and Wang, Kunyu and Wang, Shaoan and Li, Minghan and Liu, Haoran and Wei, Songlin and Wang, Zhongyuan and Zhang, Zhizheng and Wang, He},
  journal = {Robotics: Science and Systems},
  year    = {2025}
}

@article{bresenham,
  author  = {Bresenham, J. E.},
  title   = {Algorithm for Computer Control of a Digital Plotter},
  journal = {IBM Systems Journal},
  volume  = {4},
  number  = {1},
  pages   = {25--30},
  year    = {1965},
  doi     = {10.1147/sj.41.0025}
}

@inproceedings{anderson2018vision,
  title     = {Vision-and-Language Navigation: Interpreting Visually-Grounded Navigation Instructions in Real Environments},
  author    = {Anderson, Peter and Wu, Qi and Teney, Damien and Bruce, Jake and Johnson, Mark and S{\"u}nderhauf, Niko and Reid, Ian and Gould, Stephen and Van Den Hengel, Anton},
  booktitle = {Proceedings of the IEEE Conference on Computer Vision and Pattern Recognition},
  pages     = {3674--3683},
  year      = {2018}
}

@inproceedings{Jain2019StayOT,
  title     = {Stay on the Path: Instruction Fidelity in Vision-and-Language Navigation},
  author    = {Jain, Vihan and Ilharco, Gabriel and Ku, Alexander and Vaswani, Ashish and Ie, Eugene and Baldridge, Jason},
  booktitle = {Annual Meeting of the Association for Computational Linguistics},
  year      = {2019}
}

@inproceedings{krantz2020beyond,
  title        = {Beyond the Nav-Graph: Vision-and-Language Navigation in Continuous Environments},
  author       = {Krantz, Jacob and Wijmans, Erik and Majumdar, Arjun and Batra, Dhruv and Lee, Stefan},
  booktitle    = {European Conference on Computer Vision},
  pages        = {104--120},
  year         = {2020},
  organization = {Springer}
}

@article{kerbl3Dgaussians,
  author  = {Kerbl, Bernhard and Kopanas, Georgios and Leimk{\"u}hler, Thomas and Drettakis, George},
  title   = {3D Gaussian Splatting for Real-Time Radiance Field Rendering},
  journal = {ACM Transactions on Graphics},
  volume  = {42},
  number  = {4},
  month   = jul,
  year    = {2023}
}

@article{wang2025rethinking,
  title   = {Rethinking the Embodied Gap in Vision-and-Language Navigation: A Holistic Study of Physical and Visual Disparities},
  author  = {Wang, Liuyi and Xia, Xinyuan and Zhao, Hui and Wang, Hanqing and Wang, Tai and Chen, Yilun and Liu, Chengju and Chen, Qijun and Pang, Jiangmiao},
  journal = {arXiv preprint arXiv:2507.13019},
  year    = {2025}
}

@article{Qwen3-VL,
  title   = {Qwen3-VL Technical Report},
  author  = {Bai, Shuai and Cai, Yuxuan and Chen, Ruizhe and Chen, Keqin and Chen, Xionghui and Cheng, Zesen and Deng, Lianghao and Ding, Wei and Gao, Chang and Ge, Chunjiang and Ge, Wenbin and Guo, Zhifang and Huang, Qidong and Huang, Jie and Huang, Fei and Hui, Binyuan and Jiang, Shutong and Li, Zhaohai and Li, Mingsheng and Li, Mei and Li, Kaixin and Lin, Zicheng and Lin, Junyang and Liu, Xuejing and Liu, Jiawei and Liu, Chenglong and Liu, Yang and Liu, Dayiheng and Liu, Shixuan and Lu, Dunjie and Luo, Ruilin and Lv, Chenxu and Men, Rui and Meng, Lingchen and Ren, Xuancheng and Ren, Xingzhang and Song, Sibo and Sun, Yuchong and Tang, Jun and Tu, Jianhong and Wan, Jianqiang and Wang, Peng and Wang, Pengfei and Wang, Qiuyue and Wang, Yuxuan and Xie, Tianbao and Xu, Yiheng and Xu, Haiyang and Xu, Jin and Yang, Zhibo and Yang, Mingkun and Yang, Jianxin and Yang, An and Yu, Bowen and Zhang, Fei and Zhang, Hang and Zhang, Xi and Zheng, Bo and Zhong, Humen and Zhou, Jingren and Zhou, Fan and Zhou, Jing and Zhu, Yuanzhi and Zhu, Ke},
  journal = {arXiv preprint arXiv:2511.21631},
  year    = {2025}
}

@article{worldgrow2025,
  title   = {WorldGrow: Generating Infinite 3D World},
  author  = {Li, Sikuang and Yang, Chen and Fang, Jiemin and Yi, Taoran and Lu, Jia and Cen, Jiazhong and Xie, Lingxi and Shen, Wei and Tian, Qi},
  journal = {arXiv preprint arXiv:2510.21682},
  year    = {2025}
}

@inproceedings{yang2023bevformer,
  title     = {BEVFormer v2: Adapting Modern Image Backbones to Bird's-Eye-View Recognition via Perspective Supervision},
  author    = {Yang, Chenyu and Chen, Yuntao and Tian, Hao and Tao, Chenxin and Zhu, Xizhou and Zhang, Zhaoxiang and Huang, Gao and Li, Hongyang and Qiao, Yu and Lu, Lewei and others},
  booktitle = {Proceedings of the IEEE/CVF Conference on Computer Vision and Pattern Recognition},
  pages     = {17830--17839},
  year      = {2023}
}

@inproceedings{philion2020lift,
  title        = {Lift, Splat, Shoot: Encoding Images from Arbitrary Camera Rigs by Implicitly Unprojecting to 3D},
  author       = {Philion, Jonah and Fidler, Sanja},
  booktitle    = {European Conference on Computer Vision},
  pages        = {194--210},
  year         = {2020},
  organization = {Springer}
}

@inproceedings{chen2024mapgpt,
  title     = {MapGPT: Map-Guided Prompting with Adaptive Path Planning for Vision-and-Language Navigation},
  author    = {Chen, Jiaqi and Lin, Bingqian and Xu, Ran and Chai, Zhenhua and Liang, Xiaodan and Wong, Kwan-Yee K.},
  booktitle = {Proceedings of the 62nd Annual Meeting of the Association for Computational Linguistics},
  year      = {2024}
}

@article{dong2025ha,
  title   = {HA-VLN: A Benchmark for Human-Aware Navigation in Discrete-Continuous Environments with Dynamic Multi-Human Interactions, Real-World Validation, and an Open Leaderboard},
  author  = {Dong, Yifei and Wu, Fengyi and He, Qi and Li, Heng and Li, Minghan and Cheng, Zebang and Zhou, Yuxuan and Sun, Jingdong and Dai, Qi and Cheng, Zhi-Qi and others},
  journal = {arXiv preprint},
  year    = {2025}
}

@misc{liu2026navgsimhighfidelitygaussiansplatting,
  title         = {NavGSim: High-Fidelity Gaussian Splatting Simulator for Large-Scale Navigation},
  author        = {Liu, Jiahang and Duan, Yuanxing and Zhang, Jiazhao and Li, Minghan and Wang, Shaoan and Zhang, Zhizheng and Wang, He},
  year          = {2026},
  eprint        = {2603.15186},
  archivePrefix = {arXiv}
}

@inproceedings{schoenberger2016vote,
  author    = {Sch\"{o}nberger, Johannes Lutz and Price, True and Sattler, Torsten and Frahm, Jan-Michael and Pollefeys, Marc},
  title     = {A Vote-and-Verify Strategy for Fast Spatial Verification in Image Retrieval},
  booktitle = {Asian Conference on Computer Vision (ACCV)},
  year      = {2016}
}

@misc{InteriorGS2025,
  title        = {InteriorGS: A 3D Gaussian Splatting Dataset of Semantically Labeled Indoor Scenes},
  author       = {SpatialVerse Research Team, Manycore Tech Inc.},
  year         = {2025},
  howpublished = {\url{https://huggingface.co/datasets/spatialverse/InteriorGS}}
}

@article{RICHALET1978413,
  title   = {Model Predictive Heuristic Control: Applications to Industrial Processes},
  author  = {Richalet, J. and Rault, A. and Testud, J. L. and Papon, J.},
  journal = {Automatica},
  volume  = {14},
  number  = {5},
  pages   = {413--428},
  year    = {1978}
}

@misc{chen2026socialnavtraininghumaninspiredfoundation,
  title         = {SocialNav: Training Human-Inspired Foundation Model for Socially-Aware Embodied Navigation},
  author        = {Chen, Ziyi and Guo, Yingnan and Chu, Zedong and Luo, Minghua and Shen, Yanfen and Sun, Mingchao and Hu, Junjun and Xie, Shichao and Yang, Kuan and Shi, Pei and Gu, Zhining and Liu, Lu and Han, Honglin and Wu, Xiaolong and Xu, Mu and Zhang, Yu and Guo, Ning},
  year          = {2026},
  eprint        = {2511.21135},
  archivePrefix = {arXiv}
}

@misc{shao2024deepseekmathpushinglimitsmathematical,
  title         = {DeepSeekMath: Pushing the Limits of Mathematical Reasoning in Open Language Models},
  author        = {Shao, Zhihong and Wang, Peiyi and Zhu, Qihao and Xu, Runxin and Song, Junxiao and Bi, Xiao and Zhang, Haowei and Zhang, Mingchuan and Li, Y. K. and Wu, Y. and Guo, Daya},
  year          = {2024},
  eprint        = {2402.03300},
  archivePrefix = {arXiv}
}

@article{4082128,
  author  = {Hart, Peter E. and Nilsson, Nils J. and Raphael, Bertram},
  title   = {A Formal Basis for the Heuristic Determination of Minimum Cost Paths},
  journal = {IEEE Transactions on Systems Science and Cybernetics},
  volume  = {4},
  number  = {2},
  pages   = {100--107},
  year    = {1968}
}

@article{anderson2018evaluation,
  title   = {On Evaluation of Embodied Navigation Agents},
  author  = {Anderson, Peter and Chang, Angel and Chaplot, Devendra Singh and Dosovitskiy, Alexey and Gupta, Saurabh and Koltun, Vladlen and Kosecka, Jana and Malik, Jitendra and Mottaghi, Roozbeh and Savva, Manolis and others},
  journal = {arXiv preprint arXiv:1807.06757},
  year    = {2018}
}

@inproceedings{song2025towards,
  title     = {Towards Long-Horizon Vision-Language Navigation: Platform, Benchmark and Method},
  author    = {Song, Xinshuai and Chen, Weixing and Liu, Yang and Chen, Weikai and Li, Guanbin and Lin, Liang},
  booktitle = {Proceedings of the IEEE/CVF Conference on Computer Vision and Pattern Recognition},
  pages     = {12078--12088},
  year      = {2025}
}

@article{an2024etpnav,
  title   = {ETPNav: Evolving Topological Planning for Vision-Language Navigation in Continuous Environments},
  author  = {An, Dong and Wang, Hanqing and Wang, Wenguan and Wang, Zun and Huang, Yan and He, Keji and Wang, Liang},
  journal = {IEEE Transactions on Pattern Analysis and Machine Intelligence},
  year    = {2024}
}

@misc{chabal2025fomnavfrontierobjectmapsobject,
  title         = {FOM-Nav: Frontier-Object Maps for Object Goal Navigation},
  author        = {Chabal, Thomas and Chen, Shizhe and Ponce, Jean and Schmid, Cordelia},
  year          = {2025},
  eprint        = {2512.01009},
  archivePrefix = {arXiv},
  primaryClass  = {cs.RO},
  url           = {https://arxiv.org/abs/2512.01009}
}

@inproceedings{yang20253d,
  title     = {3D-Mem: 3D Scene Memory for Embodied Exploration and Reasoning},
  author    = {Yang, Yuncong and Yang, Han and Zhou, Jiachen and Chen, Peihao and Zhang, Hongxin and Du, Yilun and Gan, Chuang},
  booktitle = {Proceedings of the IEEE/CVF Conference on Computer Vision and Pattern Recognition},
  pages     = {17294--17303},
  year      = {2025}
}

@inproceedings{huang2025cogddn,
  title={Cogddn: A cognitive demand-driven navigation with decision optimization and dual-process thinking},
  author={Huang, Yuehao and Liu, Liang and Lei, Shuangming and Ma, Yukai and Su, Hao and Mei, Jianbiao and Zhao, Pengxiang and Gu, Yaqing and Liu, Yong and Lv, Jiajun},
  booktitle={Proceedings of the 33rd ACM International Conference on Multimedia},
  pages={5237--5246},
  year={2025}
}

@inproceedings{huang20242d,
  title={2d gaussian splatting for geometrically accurate radiance fields},
  author={Huang, Binbin and Yu, Zehao and Chen, Anpei and Geiger, Andreas and Gao, Shenghua},
  booktitle={ACM SIGGRAPH 2024 conference papers},
  pages={1--11},
  year={2024}
}

@article{chen2024pgsr,
  title={Pgsr: Planar-based gaussian splatting for efficient and high-fidelity surface reconstruction},
  author={Chen, Danpeng and Li, Hai and Ye, Weicai and Wang, Yifan and Xie, Weijian and Zhai, Shangjin and Wang, Nan and Liu, Haomin and Bao, Hujun and Zhang, Guofeng},
  journal={IEEE Transactions on Visualization and Computer Graphics},
  volume={31},
  number={9},
  pages={6100--6111},
  year={2024},
  publisher={IEEE}
}

@article{poole2022dreamfusion,
  title={Dreamfusion: Text-to-3d using 2d diffusion},
  author={Poole, Ben and Jain, Ajay and Barron, Jonathan T and Mildenhall, Ben},
  journal={arXiv preprint arXiv:2209.14988},
  year={2022}
}

@article{li2023gaussiandiffusion,
  title={Gaussiandiffusion: 3d gaussian splatting for denoising diffusion probabilistic models with structured noise},
  author={Li, Xinhai and Wang, Huaibin and Tseng, Kuo-Kun},
  journal={arXiv preprint arXiv:2311.11221},
  year={2023}
}

@article{zhao2025hunyuan3d,
  title={Hunyuan3d 2.0: Scaling diffusion models for high resolution textured 3d assets generation},
  author={Zhao, Zibo and Lai, Zeqiang and Lin, Qingxiang and Zhao, Yunfei and Liu, Haolin and Yang, Shuhui and Feng, Yifei and Yang, Mingxin and Zhang, Sheng and Yang, Xianghui and others},
  journal={arXiv preprint arXiv:2501.12202},
  year={2025}
}

@inproceedings{xiang2025structured,
  title={Structured 3d latents for scalable and versatile 3d generation},
  author={Xiang, Jianfeng and Lv, Zelong and Xu, Sicheng and Deng, Yu and Wang, Ruicheng and Zhang, Bowen and Chen, Dong and Tong, Xin and Yang, Jiaolong},
  booktitle={Proceedings of the IEEE/CVF conference on computer vision and pattern recognition},
  pages={21469--21480},
  year={2025}
}

@article{khanna2023hssd,
    author={{Khanna*}, Mukul and {Mao*}, Yongsen and Jiang, Hanxiao and Haresh, Sanjay and Shacklett, Brennan and Batra, Dhruv and Clegg, Alexander and Undersander, Eric and Chang, Angel X. and Savva, Manolis},
    title={{Habitat Synthetic Scenes Dataset (HSSD-200): An Analysis of 3D Scene Scale and Realism Tradeoffs for ObjectGoal Navigation}},
    journal={arXiv preprint},
    year={2023},
    eprint={2306.11290},
    archivePrefix={arXiv},
    primaryClass={cs.CV}
}

@inproceedings{lou2025robo,
  title={Robo-gs: A physics consistent spatial-temporal model for robotic arm with hybrid representation},
  author={Lou, Haozhe and Liu, Yurong and Pan, Yike and Geng, Yiran and Chen, Jianteng and Ma, Wenlong and Li, Chenglong and Wang, Lin and Feng, Hengzhen and Shi, Lu and others},
  booktitle={2025 IEEE International Conference on Robotics and Automation (ICRA)},
  pages={15379--15386},
  year={2025},
  organization={IEEE}
}

@article{li2024robogsim,
  title={Robogsim: A real2sim2real robotic gaussian splatting simulator},
  author={Li, Xinhai and Li, Jialin and Zhang, Ziheng and Zhang, Rui and Jia, Fan and Wang, Tiancai and Fan, Haoqiang and Tseng, Kuo-Kun and Wang, Ruiping},
  journal={arXiv preprint arXiv:2411.11839},
  year={2024}
}

@inproceedings{qureshi2025splatsim,
  title={Splatsim: Zero-shot sim2real transfer of rgb manipulation policies using gaussian splatting},
  author={Qureshi, M Nomaan and Garg, Sparsh and Yandun, Francisco and Held, David and Kantor, George and Silwal, Abhisesh},
  booktitle={2025 IEEE International Conference on Robotics and Automation (ICRA)},
  pages={6502--6509},
  year={2025},
  organization={IEEE}
}

@article{lei2025gaussnav,
  title={Gaussnav: Gaussian splatting for visual navigation},
  author={Lei, Xiaohan and Wang, Min and Zhou, Wengang and Li, Houqiang},
  journal={IEEE Transactions on Pattern Analysis and Machine Intelligence},
  volume={47},
  number={5},
  pages={4108--4121},
  year={2025},
  publisher={IEEE}
}

@article{wang2025trackvla,
  title={Trackvla: Embodied visual tracking in the wild},
  author={Wang, Shaoan and Zhang, Jiazhao and Li, Minghan and Liu, Jiahang and Li, Anqi and Wu, Kui and Zhong, Fangwei and Yu, Junzhi and Zhang, Zhizheng and Wang, He},
  journal={arXiv preprint arXiv:2505.23189},
  year={2025}
}

@inproceedings{cao2025cognav,
  title={Cognav: Cognitive process modeling for object goal navigation with llms},
  author={Cao, Yihan and Zhang, Jiazhao and Yu, Zhinan and Liu, Shuzhen and Qin, Zheng and Zou, Qin and Du, Bo and Xu, Kai},
  booktitle={Proceedings of the IEEE/CVF International Conference on Computer Vision},
  pages={9550--9560},
  year={2025}
}

@article{gao2025octonav,
  title={Octonav: Towards generalist embodied navigation},
  author={Gao, Chen and Jin, Liankai and Peng, Xingyu and Zhang, Jiazhao and Deng, Yue and Li, Annan and Wang, He and Liu, Si},
  journal={arXiv preprint arXiv:2506.09839},
  year={2025}
}

@article{xu2025mm,
  title={MM-Nav: Multi-View VLA Model for Robust Visual Navigation via Multi-Expert Learning},
  author={Xu, Tianyu and Chen, Jiawei and Zhang, Jiazhao and Zhang, Wenyao and Qi, Zekun and Li, Minghan and Zhang, Zhizheng and Wang, He},
  journal={arXiv preprint arXiv:2510.03142},
  year={2025}
}

@article{liu2025trackvla++,
  title={Trackvla++: Unleashing reasoning and memory capabilities in vla models for embodied visual tracking},
  author={Liu, Jiahang and Qi, Yunpeng and Zhang, Jiazhao and Li, Minghan and Wang, Shaoan and Wu, Kui and Ye, Hanjing and Zhang, Hong and Chen, Zhibo and Zhong, Fangwei and others},
  journal={arXiv preprint arXiv:2510.07134},
  year={2025}
}

@inproceedings{ross2011reduction,
  title={A reduction of imitation learning and structured prediction to no-regret online learning},
  author={Ross, St{\'e}phane and Gordon, Geoffrey and Bagnell, Drew},
  booktitle={Proceedings of the fourteenth international conference on artificial intelligence and statistics},
  pages={627--635},
  year={2011},
  organization={JMLR Workshop and Conference Proceedings}
}

@article{Chen2022ThinkGA,
  title={Think Global, Act Local: Dual-scale Graph Transformer for Vision-and-Language Navigation},
  author={Shizhe Chen and Pierre-Louis Guhur and Makarand Tapaswi and Cordelia Schmid and Ivan Laptev},
  journal={2022 IEEE/CVF Conference on Computer Vision and Pattern Recognition (CVPR)},
  year={2022},
  pages={16516-16526}
}

@inproceedings{Zhou2023NavGPTER,
  title={NavGPT: Explicit Reasoning in Vision-and-Language Navigation with Large Language Models},
  author={Gengze Zhou and Yicong Hong and Qi Wu},
  booktitle={AAAI Conference on Artificial Intelligence},
  year={2023}
}

@inproceedings{zheng2024towards,
  title={Towards learning a generalist model for embodied navigation},
  author={Zheng, Duo and Huang, Shijia and Zhao, Lin and Zhong, Yiwu and Wang, Liwei},
  booktitle={Proceedings of the IEEE/CVF Conference on Computer Vision and Pattern Recognition},
  pages={13624--13634},
  year={2024}
}

@inproceedings{chen2022pix2seq,
title     = {Pix2Seq: A Language Modeling Framework for Object Detection},
author    = {Chen, Ting and Saxena, Saurabh and Li, Lala and Fleet, David J. and Hinton, Geoffrey},
booktitle = {International Conference on Learning Representations},
year      = {2022}
}

@inproceedings{chen2022unified,
title     = {A Unified Sequence Interface for Vision Tasks},
author    = {Chen, Ting and Saxena, Saurabh and Li, Lala and Lin, Tsung-Yi and Fleet, David J. and Hinton, Geoffrey},
booktitle = {Advances in Neural Information Processing Systems},
year      = {2022}
}

@article{zhou2026deconav,
  title={DeCoNav: Dialog enhanced Long-Horizon Collaborative Vision-Language Navigation},
  author={Zhou, Sunyao and Wu, Yunzi and Wang, Tianhang and Li, Xinhai and Chen, Guang and Liu, Lizheng and Bai, Chenjia and Li, Xuelong},
  journal={arXiv preprint arXiv:2604.12486},
  year={2026}
}

@article{jing2026assemlm,
  title={AssemLM: Spatial Reasoning Multimodal Large Language Models for Robotic Assembly},
  author={Jing, Zhi and Qiao, Jinbin and Lu, Ouyang and Ao, Jicong and Qiu, Shuang and Jiang, Yu-Gang and Bai, Chenjia},
  journal={arXiv preprint arXiv:2604.08983},
  year={2026}
}

@article{an2026ai,
  title={Ai flow: Perspectives, scenarios, and approaches},
  author={An, Hongjun and Hu, Wenhan and Huang, Sida and Huang, Siqi and Li, Ruanjun and Liang, Yuanzhi and Shao, Jiawei and Song, Yiliang and Wang, Zihan and Yuan, Cheng and others},
  journal={Vicinagearth},
  volume={3},
  number={1},
  pages={1},
  year={2026},
  publisher={Springer}
}

@inproceedings{peng2024kosmos2,
title     = {Kosmos-2: Grounding Multimodal Large Language Models to the World},
author    = {Peng, Zhiliang and Wang, Wenhui and Dong, Li and Hao, Yaru and Huang, Shaohan and Ma, Shuming and Ye, Qixiang and Wei, Furu},
booktitle = {International Conference on Learning Representations},
year      = {2024}
}

@misc{jiang2025rexomni,
title         = {Detect Anything via Next Point Prediction},
author        = {Jiang, Qing and Huo, Junan and Chen, Xingyu and Xiong, Yuda and Zeng, Zhaoyang and Chen, Yihao and Ren, Tianhe and Yu, Junzhi and Zhang, Lei},
year          = {2025},
eprint        = {2510.12798},
archivePrefix = {arXiv},
primaryClass  = {cs.CV}
}

@misc{schulman2017proximal,
title         = {Proximal Policy Optimization Algorithms},
author        = {Schulman, John and Wolski, Filip and Dhariwal, Prafulla and Radford, Alec and Klimov, Oleg},
year          = {2017},
eprint        = {1707.06347},
archivePrefix = {arXiv},
primaryClass  = {cs.LG}
}

@misc{yu2025dapo,
title         = {DAPO: An Open-Source LLM Reinforcement Learning System at Scale},
author        = {Yu, Qiying and Zhang, Zheng and Zhu, Ruofei and Yuan, Yufeng and Zuo, Xiaochen and Yue, Yu and Dai, Weinan and Fan, Tiantian and Liu, Gaohong and Liu, Lingjun and Liu, Xin and Lin, Haibin and Lin, Zhiqi and Ma, Bole and Sheng, Guangming and Tong, Yuxuan and Zhang, Chi and Zhang, Mofan and Zhang, Wang and Zhu, Hang and Zhu, Jinhua and Chen, Jiaze and Chen, Jiangjie and Wang, Chengyi and Yu, Hongli and Song, Yuxuan and Wei, Xiangpeng and Zhou, Hao and Liu, Jingjing and Ma, Wei-Ying and Zhang, Ya-Qin and Yan, Lin and Qiao, Mu and Wu, Yonghui and Wang, Mingxuan},
year          = {2025},
eprint        = {2503.14476},
archivePrefix = {arXiv},
primaryClass  = {cs.LG}
}

@inproceedings{ho2020denoising,
title     = {Denoising Diffusion Probabilistic Models},
author    = {Ho, Jonathan and Jain, Ajay and Abbeel, Pieter},
booktitle = {Advances in Neural Information Processing Systems},
volume    = {33},
pages     = {6840--6851},
year      = {2020}
}

@inproceedings{chi2023diffusionpolicy,
title     = {Diffusion Policy: Visuomotor Policy Learning via Action Diffusion},
author    = {Chi, Cheng and Xu, Zhenjia and Feng, Siyuan and Cousineau, Eric and Du, Yilun and Burchfiel, Benjamin and Tedrake, Russ and Song, Shuran},
booktitle = {Robotics: Science and Systems},
year      = {2023}
}

\end{document}